%% file: main.tex
\documentclass{article} % For LaTeX2e
\usepackage{iclr2027_conference,times}

\input{math_commands.tex}

\usepackage{booktabs}
\usepackage{hyperref}
\usepackage{url}
\usepackage{multirow}
\usepackage[nameinlink,capitalise]{cleveref}
\usepackage{enumitem}
\usepackage[table,dvipsnames]{xcolor}
\setlist[enumerate]{noitemsep, topsep=0pt, parsep=0pt, partopsep=0pt}
\usepackage{graphicx}
\usepackage{float}
\usepackage{xcolor}
\usepackage{subcaption}
\usepackage{tabularx}
\usepackage{float}
\usepackage{algorithm}
\usepackage{algorithmicx}
\usepackage{algpseudocode}
\usepackage{duckuments}
\usepackage{pgfplots}
\usepackage{amssymb}
\usepackage{subcaption}

\usepackage[table]{xcolor}

\definecolor{indomain}{RGB}{242,242,242}
\definecolor{transfer}{RGB}{225,238,248}
\definecolor{oodcell}{RGB}{252,239,218}
\definecolor{overall}{RGB}{242,246,249}
\definecolor{jointrow}{RGB}{228,238,246}

\usepackage{pifont}

\usepackage{pgfplots}
\definecolor{mycitecolor}{RGB}{0, 102, 204}  % You can customize this

\usepackage{listings}
\definecolor{codekeyword}{rgb}{0.74, 0.0, 0.44}   % magenta/pink for def, for, return
\definecolor{codecomment}{rgb}{0.0, 0.5, 0.5}     % teal for comments
\lstdefinestyle{pytorch}{
    commentstyle=\color{codecomment},
    keywordstyle=\color{codekeyword}\bfseries,
    basicstyle=\ttfamily\small,
    breaklines=true,
    columns=fullflexible,
    keepspaces=true,
    showstringspaces=false,
    language=Python,
    morekeywords={def, for, in, range, return},
    frame=none,
    xleftmargin=0pt,
    xrightmargin=0pt,
}

\title{DIVINE: Simple Cross-Market Stock Pretraining via Diverse Indicator Reconstruction}

\author{%
  \textbf{Kuan-Yu Chen}$^{*}$\textbf{,}
  \textbf{Shu-Cheng Zheng,}
  \textbf{Yu-Chen Den,}
  \textbf{Wei-Cheng Liao}$^{\dagger}$\textbf{\&}
  \textbf{Tien-Hao Chang} \\
  SinoPac Holdings, Taipei, Taiwan \\
  \small\texttt{\{lavamore,samuelzheng,abnerden,wcliao,darby\}@sinopac.com
  }
}

\newcommand{\model}[1]{{DIVINE}}
\iclrfinalcopy

\begin{document}

\maketitle

\renewcommand{\thefootnote}{}
\footnotetext{$^*$Corresponding Author.}
\footnotetext{$^\dagger$Work performed while at SinoPac Holdings.}
\renewcommand{\thefootnote}{\arabic{footnote}}

\begin{abstract}
Financial time-series pretraining typically learns from masked observations, contrastive relations, or future outcomes---yet existing objectives struggle to simultaneously avoid future-supervision uncertainty and maintain return-prediction alignment.
We propose DIVINE (DIVerse INdicator rEconstruction), a simple cross-market pretraining framework that reconstructs technical indicators from raw OHLCV history.
Computed from observed price-volume history, technical indicators provide consistently defined supervision across markets while summarizing diverse market dynamics with established relevance to return prediction.
Pretrained jointly on six-equity market datasets, DIVINE reconstructs 77 targets derived from 16 standard indicators and transfers only the learned encoder to downstream stock ranking. Across all six markets, DIVINE achieves the strongest average portfolio performance with a lightweight 0.05M-parameter encoder, outperforming pretraining baselines and matching or exceeding substantially larger financial foundation models, while remaining robust and data-efficient.
Systematic analyses show that indicator diversity and market diversity provide complementary gains in transfer. 
Together, these results suggest that supervision design and cross-market diversity—rather than model scale—are the key drivers of strong, transferable financial representations.
\end{abstract}

\input{content/introduction}
\input{content/related_work}
\input{content/preliminary}
\input{content/method}
\input{content/main_experiment}

\input{content/ablation}
\input{content/robuestness_efficiency}
\input{content/analysis}
\input{content/conclusion}

\section*{AI Use Statement}
We used generative AI tools for literature survey, feedback on experimental validation, and assistance with interpreting experimental results. We additionally used generative AI for manuscript organization, drafting assistance, and language editing. All scientific claims, experimental results, retrieved references, and AI-assisted content were critically reviewed and verified by the authors. The authors take full responsibility for the final content of this work.

\section*{Reproducibility Statement}
To support reproducibility, we provide detailed specifications of the experimental pipeline throughout the main paper and appendices.
Appendix~\ref{app:indicator} specifies the technical-indicator targets and their configurations;
Appendix~\ref{app:data_and_eval} describes data acquisition, stock-universe construction, temporal splits, input and target construction, and the
portfolio backtesting protocol; 
Appendix~\ref{app:hydra_implementation} provides the target preprocessing procedure together with model architectures, training configurations, and optimization settings of \model{}; 
and Appendix~\ref{app:baseline_implementation} details the implementation, hyperparameters, and transfer protocols of the compared baselines.

\bibliography{reference}
\bibliographystyle{iclr2027_conference}

\clearpage
\appendix
\input{content/appendix_indicator}

\end{document}

%% file: math_commands.tex
\usepackage{amsmath,amsfonts,bm}

\def\eqref#1{equation~\ref{#1}}
\def\1{\bm{1}}

\DeclareMathAlphabet{\mathsfit}{\encodingdefault}{\sfdefault}{m}{sl}
\SetMathAlphabet{\mathsfit}{bold}{\encodingdefault}{\sfdefault}{bx}{n}

%% file: content/introduction.tex
\section{Introduction}

Stock return prediction is a long-standing challenge in quantitative finance, particularly in cross-sectional settings where models must rank assets by expected future returns under highly non-stationary market dynamics~\citep{azariadis1998financial, avramov2002stock, hu2025fintsb} and low signal-to-noise conditions~\citep{rapach2010out, gu2020empirical, caner2025deep}.
A key difficulty is that equity markets differ in their cross-sectional and temporal structure, which strongly shapes return predictability~\citep{feng2019temporal, hsu2021fingat, li2024master}.
Models trained on a single market may therefore specialize to its market-specific regularities, limiting their ability to transfer across regimes and markets.
Learning reusable representations across markets is thus appealing: it can capture recurring price-volume structure across financial systems while reducing dependence on labeled data from any single market.

Pretraining offers a natural route toward such transferable representations, but its effectiveness depends critically on the supervision objective.
Existing approaches largely reconstruct masked prices~\citep{he2022mae, nie2022time, xia2024cisthpan, chen2025mera, wang2025sspt}, learn contrastive relations between sequences~\citep{yue2022ts2vec, tian2024timesurl, lee2026cross}, predict future prices autoregressively~\citep{ding2025delphyne, zhu2025fincast, shi2025kronos}, or solve stock-specific pretext tasks~\citep{wang2025sspt}.
While effective, existing objectives often expose a tension between \emph{future-supervision uncertainty} and \emph{return-prediction alignment}.
Objectives based on future prices or returns are naturally aligned with prediction, but their targets depend on \emph{stochastic} future market realizations.
By contrast, objectives constructed solely from past observations avoid this source of uncertainty, 
but offer no explicit link between the pretraining objective and future-return prediction.
This raises a central question: \emph{\textbf{what form of past-only supervision can be directly aligned with future-return prediction?}}

We explore technical indicators as such a supervision signal.
Computed purely from past price-volume observations, they summarize diverse aspects of market dynamics---including trend, volatility, momentum, relative price positioning, directional movement, and price-volume interaction~\citep{wilder1978, lambert1980, murphy1999study, appel2005}---with definitions that remain consistent across markets. Such price-volume transformations have long served as building blocks for predictive alpha factors in quantitative finance~\citep{kakushadze2016101, yang2020qlib, fang2020nnafc, https://doi.org/10.1111/jofi.12883}, grounding their relevance to future-return prediction and making them natural candidates for return-aligned supervision.
We therefore ask whether reconstructing a diverse indicator space can exploit this predictive structure to learn representations that transfer beyond the indicator targets themselves.

We propose \textbf{\model{}} (DIVerse INdicator rEconstruction), a simple cross-market pretraining framework using diverse technical-indicator reconstruction.
Given raw OHLCV sequences, \model{} jointly reconstructs 77 targets derived from 16 standard technical indicators across multiple period configurations.
All targets are reconstructed from a shared latent representation, encouraging the encoder to capture price-volume structure that supports diverse market summaries.
We jointly pretrain the encoder on six international equity markets, discard the reconstruction head, and transfer only the learned encoder to cross-sectional return ranking in each target market.
Technical indicators are thus used only during pretraining and are not required as downstream inputs.
We summarize our contributions as follows:
\begin{itemize}[leftmargin=*]

\item \textbf{Technical-indicator reconstruction for representation learning.}
We introduce \model{}, a cross-market pretraining framework that reconstructs a diverse space of technical indicators computed from past OHLCV history to learn transferable financial representations. Rather than using technical indicators directly as downstream features, \model{} uses them only as pretraining targets to shape a latent representation that transfers to stock ranking.

\item \textbf{Strong performance with lightweight and data-efficient pretraining.}
Across six equity markets, \model{} achieves the strongest average portfolio performance with a lightweight 0.05M-parameter encoder, outperforming pretraining methods spanning diverse supervision paradigms, and matching or exceeding financial foundation models pretrained on substantially larger external corpora with approximately $490\times$ fewer parameters.
It further remains robust across encoder architectures and downstream settings. Beyond strong portfolio performance, \model{} is remarkably efficient: it surpasses full-data pretraining baselines using only 10\% of the pretraining corpus, and exceeds full-data training from scratch with only 5\% of downstream labels.

\item \textbf{Indicator and market diversity jointly drive transfer.}
Through systematic ablations and transfer analyses, we show that the learned representation outperforms both direct indicator features and the reconstructed indicators themselves. 
Broader indicator type coverage contributes substantially more than adding period variants, while even pretraining without the target market still recovers 64\% of the single-market-to-joint improvement.
These results establish indicator diversity and market diversity as complementary drivers of transferable financial representations.
\end{itemize}

%% file: content/related_work.tex
\section{Related Work}
\label{sec:related_work}

\paragraph{Supervised Stock Ranking and Prediction.}
Deep models for stock ranking span recurrent architectures that optimize investment returns directly rather than price regression~\citep{feng2018enhancing,feng2019temporal}, models that exploit inter-stock structure~\citep{yoo2021accurate,hsu2021fingat,li2024master}, mixture-of-experts frameworks~\citep{sun2023alphamix,chen2025dhmoe}, and lightweight architectures that mix information across indicator, temporal, and stock dimensions~\citep{fan2024stockmixer}.
Most of these methods are trained from scratch on individual markets.
QuantNet~\citep{koshiyama2021quantnet} instead couples market-specific encoders through a shared bottleneck, but optimizes it end-to-end for trading signals rather than explicitly pretraining a transferable representation.

\paragraph{Financial Time-Series Pretraining.}
Existing pretraining methods can be broadly organized by how their supervision relates historical observations to future returns.
\textbf{(1)~Masked price reconstruction} adapts masked autoencoding~\citep{he2022mae, nie2022time} to stock sequences, including variants using dynamic hypergraphs~\citep{sawhney2021stock,xia2024cisthpan} or retrieval-augmented expert routing~\citep{chen2025mera}.
Such objectives avoid future-outcome supervision, but reconstruction of raw missing observations is only indirectly related to future-return prediction and can be sensitive to short-term variation.
SSPT~\citep{wang2025sspt} instead reconstructs moving averages over masked windows, partially mitigating such variation.
\textbf{(2)~Contrastive learning} defines positive and negative relations through temporal augmentation~\citep{yue2022ts2vec, tian2024timesurl} or cross-stock future-return correlation~\citep{lee2026cross}.
The former is only indirectly aligned with future returns, whereas the latter relies on future outcomes and thus inherits uncertainty from their stochastic realizations.
\textbf{(3)~Autoregressive prediction} predicts future prices over large multi-market or multi-asset corpora~\citep{ding2025delphyne, shi2025kronos, zhu2025fincast}.
Although naturally aligned with forecasting, its supervision is inherently uncertain because realized future prices contain stochastic variation not determined by observed history.
\textbf{(4)~Stock-specific pretext tasks} use auxiliary metadata such as stock identity and sector as targets~\citep{wang2025sspt}. Such objectives do not rely on future outcomes, but their connection to future-return prediction remains indirect.
Overall, existing objectives often trade off future-outcome independence against downstream alignment.
\model{} instead reconstructs technical indicators from observed history, avoiding future-outcome supervision while retaining established relevance to future-return prediction.

\paragraph{Technical Indicators as Supervision Signals.}
Technical indicators summarize diverse price-volume dynamics and are widely used in alpha-factor construction~\citep{kakushadze2016101, yang2020qlib}.
NNAFC~\citep{fang2020nnafc}, the closest prior work using indicators as neural supervision targets, further shows that indicator-derived expressions can carry cross-sectional predictive signal.
However, it trains a separate network for each indicator to generate tradable factors within a single market rather than learning a reusable shared representation.
In contrast, \model{} jointly reconstructs a diverse indicator space, discards the indicator predictions after pretraining, and transfers only the shared encoder representation across markets.

%% file: content/preliminary.tex
\section{Problem Formulation}
\label{sec:problem}

\paragraph{Notation.}
Let $\mathcal{M}=\{m_1,\dots,m_M\}$ denote a set of equity markets, where market $m$ contains $T_m$ trading days and a stock universe $\mathcal{S}^m$.
For stock $i\in\mathcal{S}^m$, let $\mathbf{X}^m_i\in\mathbb{R}^{T_m\times 5}$ denote its open, high, low, close, and volume (OHLCV) history. 
At day $t$, let $\mathbf{X}^m_{i,t-L+1:t}\in\mathbb{R}^{L\times 5}$ be the length-$L$ lookback window and
$r^m_{i,t\rightarrow t+h} = (c^m_{i,t+h} - c^m_{i,t}) / c^m_{i,t} \in \mathbb{R}$
the realized return over the next $h$ days, where $c^m_{i,t}$ denotes the closing price.

\paragraph{Cross-Sectional Return Ranking.}
For each market $m$, we learn a scoring function $s^m(\cdot)$ that ranks stocks from their recent OHLCV histories such that higher-scored stocks achieve higher future returns.
Following prior work~\citep{fang2019alpha, den2026integrating}, we maximize the daily Spearman rank correlation between predicted scores and realized returns:
\begin{equation}
\max_{s^m}\;
\mathbb{E}_{t}\!\left[
\rho\!\left(
\{s^m(\mathbf{X}^m_{i,t-L+1:t})\}_{i\in\mathcal{S}^m},
\{r^m_{i,t\rightarrow t+h}\}_{i\in\mathcal{S}^m}
\right)
\right].
\label{eq:objective}
\end{equation}
We optimize~\eqref{eq:objective} using a differentiable rank surrogate~\citep{fang2019alpha} and evaluate the resulting rankings through portfolio backtesting. Implementation details of backtesting and evaluation metrics are provided in Section~\ref{app:evaluation}.

\paragraph{Pretraining and Downstream Transfer.}
We parameterize the scoring function as $s^m=\phi^m\circ g_\theta$, where $g_\theta:\mathbb{R}^{L\times5}\rightarrow\mathbb{R}^{d_{\mathrm{enc}}}$ is a shared encoder and $\phi^m:\mathbb{R}^{d_{\mathrm{enc}}}\rightarrow\mathbb{R}$ is a lightweight market-specific prediction head.
The encoder is first pretrained on a corpus $\mathcal{D}_{\mathrm{pre}}$ drawn from markets $\mathcal{M}_{\mathrm{pre}}\subseteq\mathcal{M}$, then transferred to market $m$, where $\phi^m$ is optimized under~\eqref{eq:objective} with $\theta$ either frozen or jointly updated (\emph{full fine-tuning}).
We call $\mathcal{M}_{\mathrm{pre}}=\{m\}$ as \emph{single-market pretraining} and $|\mathcal{M}_{\mathrm{pre}}|>1$ as \emph{cross-market pretraining}. Additionally, when $(|\mathcal{M}_{\mathrm{pre}}|>1) \wedge (m\notin\mathcal{M}_{\mathrm{pre}})$, we refer to it as \emph{target-excluded cross-market pretraining}.
As a reference baseline, training from scratch omits pretraining and optimizes both $\theta$ and $\phi^m$ directly on the downstream objective.

%% file: content/method.tex
\section{Cross-Market Pretraining via Indicator Reconstruction}
\label{sec:method}

\begin{figure}[t]
\centering
\includegraphics[width=\textwidth]{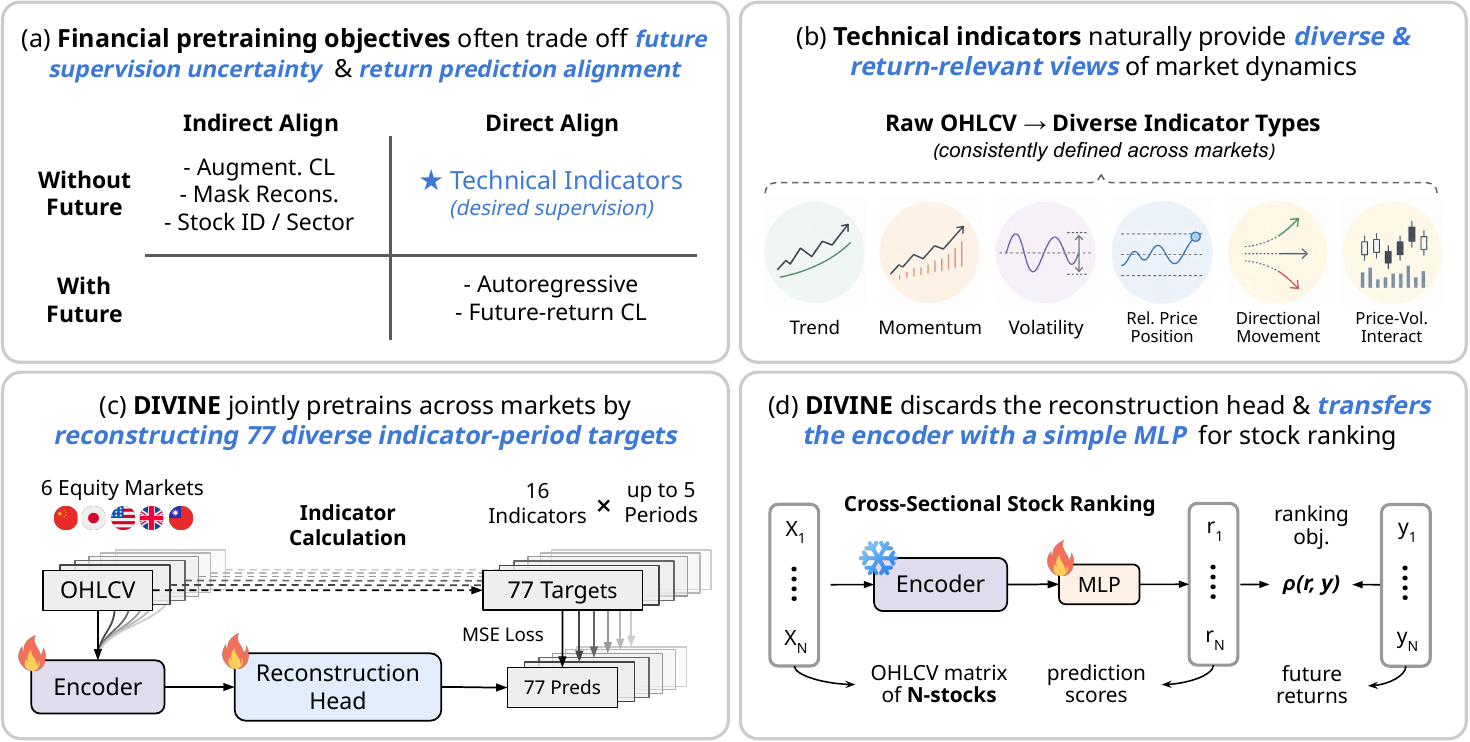}
\caption{
Overview of \model{}, which jointly reconstructs diverse technical indicators across markets and transfers the pretrained encoder to downstream stock ranking.}
\label{fig:hydra77}
\end{figure}

\model{} pretrains an encoder to reconstruct technical indicators from raw OHLCV windows.
Figure~\ref{fig:hydra77} provides an overview of \model{}, while Sections~\ref{sec:targets} and~\ref{sec:objective} detail the indicator targets and pretraining/transfer protocol, respectively.

\subsection{Diverse Technical Indicators as Pretraining Targets}
\label{sec:targets}

\paragraph{Target Definition.}
Given the observed price-volume history of stock $i$ in market $m$ up to day $t$, we define
\begin{equation}
    \mathbf{z}_{i,t}^{m}
    = \Psi(\mathbf{X}^m_{i,:t}) \in \mathbb{R}^{D},
    \label{eq:indicator_target}
\end{equation}
where $\Psi=[\psi_{k_1}^{j_1},\ldots,\psi_{k_D}^{j_D}]$ collects $D$ indicator--period combinations, with each $\psi_k^j$ denoting indicator $j$ evaluated at period $k$.
We adopt 16 standard technical indicators (e.g., SMA, RSI, MACD, ATR, and MFI) spanning diverse aspects of price-volume dynamics, including trend, volatility, momentum, relative price positioning, directional movement, and price-volume interaction.
Most indicators are computed at five periods $k \in \{3, 5, 10, 14, 20\}$, with indicator-specific exceptions, yielding $D=77$ indicator--period targets in total.
The indicator set is selected for breadth of market-dynamic coverage rather than downstream performance. The full list of indicators is provided in Appendix~\ref{app:indicator}.

\paragraph{Target Normalization.}
To prevent scale differences across indicators and markets from dominating the reconstruction loss, we normalize inputs and targets in three stages.
\emph{First}, OHLCV inputs and price- or volume-level-dependent targets are divided by the corresponding maximum price or volume within the input window, while scale-free indicators remain unchanged.
\emph{Second}, each target dimension is clipped to $\mathrm{median}\pm10\times\mathrm{IQR}$ and standardized using training-split statistics from its market.
\emph{Third}, for cross-market pretraining, targets are re-standardized using pooled statistics across all pretraining markets.
The resulting target is denoted as $\tilde{\mathbf{z}}_{i,t}^{m}\in\mathbb{R}^{D}$.
Full preprocessing details and ablations are provided in Appendices~\ref{app:preprocess_algo} and~\ref{app:preprocess_ablation}, respectively.

\subsection{Pretraining Objective and Downstream Transfer}
\label{sec:objective}

\paragraph{Indicator Reconstruction.}
Given an OHLCV window $\mathbf{X}_{i,t-L+1:t}^{m}\in\mathbb{R}^{L\times5}$, the encoder produces
$\mathbf{h}_{i,t}^{m}=g_\theta(\mathbf{X}_{i,t-L+1:t}^{m})\in\mathbb{R}^{d}$,
and a linear reconstruction head $\mathbf{W}\in\mathbb{R}^{d\times D}$ predicts all $D$ indicator targets from this shared representation, encouraging the encoder to capture price-volume structure shared across diverse indicator types and period scales.
We train the encoder and reconstruction head by minimizing the mean squared reconstruction error:
\begin{equation}
\mathcal{L}(\theta,\mathbf{W})
=
\mathbb{E}_{m,i,t}
\left[
\left\|
\mathbf{W}^{\top}g_\theta(\mathbf{X}_{i,t-L+1:t}^{m})
-
\tilde{\mathbf{z}}_{i,t}^{m}
\right\|_2^2
\right].
\label{eq:pretrain}
\end{equation}
To give all markets equal expected contribution during joint pretraining, markets are sampled inversely proportional to their training-set frequency.
The reconstruction quality of each indicator--period target is reported in Appendix~\ref{app:reconstruction}.

\paragraph{Architecture and Transfer.}
Our default encoder $g_\theta$ is a two-layer GRU whose last hidden state forms the representation $\mathbf{h}_{i,t}^{m}$, with evaluations of alternative encoder architectures provided in Section~\ref{sec:main_results}.
After pretraining, we discard the reconstruction head $\mathbf{W}$, attach a two-layer MLP $\phi^m$ for each downstream market, and optimize it using the ranking objective in~\eqref{eq:objective}.
Detailed implementation and hyperparameter settings of \model{} are provided in Appendix~\ref{app:hydra_hyperparams}.

%% file: content/main_experiment.tex
\section{Experiments}

\subsection{Experimental Setup}
\label{sec:setup}

\paragraph{Datasets and Inputs}
We evaluate \model{} on six equity-market datasets: CSI300 (CN), CSI500 (CN), NI225 (JP), SP500 (US), FTSE100 (UK), and TWSE300 (TW), with 2008--2019 for training, 2020 for validation, and 2021--2024 for testing. Detailed dataset statistics and the construction of stock universe are provided in Appendix~\ref{app:datasets}.
Following prior work~\citep{feng2019temporal, sun2023alphamix, den2026integrating}, most baselines take OHLCV concatenated with three moving-average features as input, with preprocessing details provided in Appendix~\ref{app:dataset_preprocessing}.
Exceptions are methods whose input formats are tied to their pretraining. For example, financial foundation models~\citep{shi2025kronos, zhu2025fincast} are pretrained on their released preprocessing pipelines, and \model{} takes raw OHLCV as input.

\paragraph{Compared Baselines.}
We compare \model{} against four groups of methods:
\textbf{(1)~Financial models trained from scratch}: RankLSTM~\citep{feng2019temporal}, FinGAT~\citep{hsu2021fingat}, AlphaMix~\citep{sun2023alphamix}, MASTER~\citep{li2024master}, StockMixer~\citep{fan2024stockmixer}, and DHMoE~\citep{chen2025dhmoe};
\textbf{(2)~General time-series pretraining}: TS2Vec~\citep{yue2022ts2vec}, PatchTST~\citep{nie2022time}, and TimesURL~\citep{tian2024timesurl};
\textbf{(3)~Financial pretraining}: CI-STHPAN~\citep{xia2024cisthpan}, MERA~\citep{chen2025mera}, SSPT~\citep{wang2025sspt}, and FASCL~\citep{lee2026cross}; and
\textbf{(4)~Financial foundation models}: Kronos~\citep{shi2025kronos} and FinCast~\citep{zhu2025fincast}.
Note that the released foundation models are pretrained on substantially larger external corpora with broader temporal coverage, so we include them as external references rather than controlled comparisons, providing a benchmark against strong industry-scale models.
Implementation details and hyperparameter settings for all baselines appear in Appendix~\ref{app:baseline_implementation}.

\paragraph{Downstream Evaluation Protocol.}
From-scratch and pretraining baselines use a 20-day input window and 5-day prediction horizon for cross-sectional return ranking.
For financial foundation models, we adapt pretrained representations to the same ranking task using model-specific preprocessing and adaptation procedures, and select the configuration most aligned with the evaluation setting of other baselines, with full ablations across all configurations reported in Appendix~\ref{app:foundation_models}.
Unless otherwise specified, all methods follow the same portfolio backtesting protocol and report long-only Sharpe Ratio (SR) and Calmar Ratio (CR), excluding transaction costs. 
Each setting is evaluated over 25 seed--offset combinations (five random seeds $\times$ five staggered starting offsets), with full details provided in Appendix~\ref{app:evaluation}.
Note that our datasets use fixed stock universes shared by all methods, which may introduce survivorship effects. Absolute portfolio performance should therefore be interpreted as a controlled comparison rather than a point-in-time investable backtest.

\subsection{Main Results}
\label{sec:main_results}

Table~\ref{tab:main_results} summarizes portfolio backtest results across six equity markets. Pretraining methods generally achieve stronger average performance than models trained from scratch, although the gains vary substantially across pretraining approaches. Notably, none of the compared financial pretraining methods surpasses PatchTST on average, suggesting that financial-domain pretraining alone does not necessarily yield stronger transfer.

Among all generic and financial pretraining baselines, \model{} achieves the strongest average performance.
With a GRU encoder, it reaches an average SR/CR of 1.396/1.790, improving over the strongest pretraining baseline, PatchTST (1.007/1.202), by 39\%/49\%.
The same pattern holds across encoder families: the Transformer and Mamba variants also outperform all compared pretraining baselines on average, 
showing that \model{} is not backbone-specific and remains robust under both frozen-encoder transfer and full fine-tuning (Appendix~\ref{app:main_with_freeze}).

Among financial foundation models, performance does not scale with model size: FinCast (991.4M parameters) underperforms even models trained from scratch, while Kronos-small (24.5M parameters) provides the strongest external reference.
Kronos-small reaches an average SR/CR of 1.256/1.404, yet the default GRU-based \model{} (0.05M parameters) achieves 11.1\%/27.5\% higher SR/CR with approximately $490\times$ fewer parameters, with both improvements being statistically significant across the 25 matched seed--offset evaluations.
Results under realistic transaction costs are provided in Appendix~\ref{app:main_with_cost}, where \model{} maintains the highest average SR/CR across markets.

The main exception is FTSE100, the smallest stock universe in our evaluation, where methods that explicitly or implicitly model inter-stock relationships (e.g., AlphaMix, MERA, CI-STHPAN, FASCL) match or outperform \model{}.
This suggests that cross-stock structure may be particularly important in smaller universes, highlighting a limitation of the current per-stock encoder.
While this section focuses on the default evaluation setting, Section~\ref{sec:robustness_efficiency} further evaluates robustness across prediction horizons, portfolio strategies, and data regimes.

\input{table/main_result_short}

%% file: table/main_result_short.tex
\begin{table*}[t]
\caption{Portfolio backtest results across six equity-market datasets from 2021 to 2024. For each method, we report its parameter size in millions (M) and portfolio performance in Sharpe Ratio (SR) and Calmar Ratio (CR). Best and second-best performance per column are highlighted in \textbf{bold} and \underline{underline}, respectively. $^{*}$ denotes a statistically significant improvement over Kronos-small in the average performance across markets ($p<0.05$, one-sided paired $t$-test over 25 matched seed--offset evaluations). Standard deviations of the average SR and CR results are reported in Appendix~\ref{app:main_with_std}.}

\centering
\setlength{\tabcolsep}{4pt}
\resizebox{\textwidth}{!}{%
\begin{tabular}{lc|cc|cc|cc|cc|cc|cc|cc}
\toprule
\textbf{Markets} $\rightarrow$ & & \multicolumn{2}{c}{\textbf{CSI300}} & \multicolumn{2}{c}{\textbf{CSI500}} & \multicolumn{2}{c}{\textbf{NI225}} & \multicolumn{2}{c}{\textbf{SP500}} & \multicolumn{2}{c}{\textbf{FTSE100}} & \multicolumn{2}{c}{\textbf{TWSE300}} & \multicolumn{2}{c}{\textbf{Avg.}} \\
\textbf{Method} & \textbf{Param.} & SR & CR & SR & CR & SR & CR & SR & CR & SR & CR & SR & CR & SR & CR \\
\midrule

\multicolumn{16}{l}{\textit{\textbf{Financial models trained from scratch}}} \\
RankLSTM & 0.05 & 0.753 & 0.663 & 0.869 & 0.699 & 0.610 & 0.447 & 1.414 & 2.138 & 0.314 & 0.226 & 0.645 & 0.461 & 0.768 & 0.772 \\
FinGAT & 0.04 & 0.820 & 0.536 & 1.175 & 1.063 & 0.789 & 0.634 & 0.485 & 0.428 & 0.537 & 0.330 & 0.775 & 0.578 & 0.764 & 0.595 \\
MASTER & 0.7 & 0.991 & 0.879 & 1.198 & 1.044 & 0.730 & 0.616 & 1.477 & 2.398 & 0.241 & 0.143 & 0.750 & 0.619 & 0.898 & 0.950 \\
AlphaMix & 0.04 & 0.708 & 0.584 & 1.064 & 0.959 & 0.543 & 0.458 & 1.574 & 2.486 & 0.612 & 0.448 & 0.736 & 0.650 & 0.873 & 0.931 \\
StockMixer & 0.01 & 0.726 & 0.612 & 0.615 & 0.421 & 0.450 & 0.361 & 0.919 & 0.899 & 0.407 & 0.263 & 0.763 & 0.554 & 0.647 & 0.518 \\
DHMoE & 10.2 & 0.961 & 0.977 & 1.716 & \textbf{2.490} & 0.679 & 0.585 & 0.980 & 0.680 & 0.612 & 0.384 & 0.464 & 0.292 & 0.902 & 0.901 \\
\midrule

\multicolumn{16}{l}{\textit{\textbf{General Time-Series Pretraining}}} \\
TS2Vec & 0.6 & 1.014 & 0.893 & 1.522 & 1.640 & 0.780 & 0.664 & 0.952 & 1.008 & 0.493 & 0.344 & 1.140 & 0.989 & 0.983 & 0.923 \\
PatchTST & 0.6 & 0.662 & 0.533 & 1.147 & 1.113 & 0.807 & 0.653 & 1.618 & 3.117 & 0.437 & 0.334 & 1.371 & 1.465 & 1.007 & 1.202 \\
TimesURL & 0.6 & 1.022 & 0.972 & 1.302 & 1.277 & 0.576 & 0.448 & 1.057 & 1.191 & 0.412 & 0.268 & 1.117 & 0.958 & 0.914 & 0.852 \\
\midrule

\multicolumn{16}{l}{\textit{\textbf{Financial Pretraining}}} \\
MERA & 1.5 & 1.075 & 1.055 & 0.677 & 0.472 & 0.769 & 0.554 & 1.117 & 1.131 & 0.704 & 0.469 & 1.116 & 0.955 & 0.962 & 0.833 \\
CI-STHPAN & 0.5 & 0.720 & 0.518 & 0.949 & 0.646 & 0.286 & 0.261 & 1.527 & 2.075 & \underline{0.705} & \textbf{0.553} & 0.520 & 0.423 & 0.785 & 0.746 \\
FASCL & 14.2 & 0.547 & 0.485 & 0.951 & 0.881 & 0.954 & 0.742 & 1.300 & 1.713 & \textbf{0.718} & \underline{0.504} & 1.246 & 1.275 & 0.953 & 0.933 \\
SSPT & 0.4 & 0.447 & 0.404 & 0.931 & 0.724 & 1.079 & 0.896 & 1.162 & 2.035 & 0.540 & 0.362 & 1.095 & 0.938 & 0.876 & 0.893 \\
\midrule

\multicolumn{16}{l}{\textit{\textbf{Financial Foundation Models}}} \\
FinCast & 991.4 & 0.780 & 0.668 & 1.036 & 0.819 & 0.649 & 0.507 & 0.699 & 0.560 & 0.434 & 0.274 & 0.924 & 0.673 & 0.754 & 0.583 \\
Kronos-mini & 5.3 & 1.260 & 1.344 & 1.637 & 1.822 & 0.762 & 0.592 & 1.266 & 1.644 & 0.522 & 0.365 & 1.214 & 1.085 & 1.110 & 1.142 \\
Kronos-small & 24.5 & \underline{1.459} & \underline{1.516} & \underline{1.916} & \underline{2.229} & 0.903 & 0.783 & 1.523 & 2.329 & 0.524 & 0.378 & 1.209 & 1.186 & 1.256 & 1.404 \\
Kronos-base & 99.7 & 1.326 & 1.305 & 1.803 & 1.979 & 0.895 & 0.714 & 1.387 & 2.040 & 0.359 & 0.252 & 1.154 & 1.076 & 1.154 & 1.228 \\
\midrule

\multicolumn{16}{l}{\textit{\textbf{\model{} (Ours)}}} \\
GRU encoder & 0.05 & \textbf{1.520} & \textbf{1.640} & \textbf{1.982} & 2.207 & \textbf{1.245} & \textbf{1.146} & \underline{1.645} & \textbf{3.774} & 0.568 & 0.413 & \underline{1.418} & \underline{1.549} & \textbf{1.396}$^{*}$ & \textbf{1.790}$^{*}$ \\
Mamba encoder & 0.07 & 1.083 & 1.183 & 1.675 & 1.841 & 1.075 & 0.961 & \textbf{1.807} & 3.519 & 0.645 & 0.425 & 1.273 & 1.336 & 1.259 & 1.544$^{*}$ \\
TFM encoder  & 0.1 & 1.383 & 1.371 & 1.811 & 1.813 & \underline{1.193} & \underline{0.970} & 1.456 & \underline{3.586} & 0.646 & 0.446 & \textbf{1.510} & \textbf{1.835} & \underline{1.333}$^{*}$ & \underline{1.670}$^{*}$ \\

\bottomrule
\end{tabular}%
}
\label{tab:main_results}
\end{table*}

%% file: content/ablation.tex
\subsection{Ablation Study of Indicator Supervision}
\label{sec:sanity_check}

Table~\ref{tab:indicator_analysis} conducts an ablation study isolating three sources of gain in \model{}:
(i) using indicators as inputs versus reconstruction targets, 
(ii) single-market versus cross-market pretraining, and
(iii) transferring via reconstructed indicators versus the encoder hidden state.
All settings use the same indicator space and normalization, with other components matched whenever applicable.

\input{table/ablation}

Using indicators directly as input features improves average SR/CR from 0.829/0.938 (raw OHLCV) to 1.173/1.251, confirming that the indicator space carries more predictive information than raw price-volume features alone.
Single-market indicator reconstruction further reaches 1.183/1.419, showing that reconstruction-based representation learning provides additional benefit beyond direct indicator inputs.
The largest gain comes from cross-market pretraining: holding the indicator target and downstream signal fixed, expanding pretraining from one market to multiple markets raises SR/CR from 1.183/1.419 to 1.396/1.790.
Finally, under identical cross-market pretraining, the hidden representation outperforms the reconstructed indicators (1.396/1.790 vs.\ 1.333/1.567), indicating that the encoder learns ranking-relevant structure beyond the reconstructed targets themselves.

%% file: table/ablation.tex
\begin{table}[ht]
\centering
\caption{
Ablation of \model{} along three factors: indicator role, pretraining scope, and downstream signal, reported in SR and CR. The upper block reports from-scratch baselines and the lower block reports pretraining variants, with the final row corresponding to the full \model{}. Best results per column are shown in \textbf{bold}.
}
\setlength{\tabcolsep}{4pt}
\resizebox{\textwidth}{!}{%
\begin{tabular}{ccc|cc|cc|cc|cc|cc|cc|cc}
\toprule

\multicolumn{3}{c|}{\textbf{Ablation Factors}}
& \multicolumn{2}{c}{\textbf{CSI300}}
& \multicolumn{2}{c}{\textbf{CSI500}}
& \multicolumn{2}{c}{\textbf{NI225}}
& \multicolumn{2}{c}{\textbf{SP500}}
& \multicolumn{2}{c}{\textbf{FTSE100}}
& \multicolumn{2}{c}{\textbf{TWSE300}}
& \multicolumn{2}{c}{\textbf{Avg.}} \\
\cmidrule(lr){1-3}

\begin{tabular}[c]{@{}c@{}}\textbf{Indicator}\\\textbf{Role}\end{tabular} & \begin{tabular}[c]{@{}c@{}}\textbf{Pretraining}\\\textbf{Scope}\end{tabular} & \begin{tabular}[c]{@{}c@{}}\textbf{Downstream}\\\textbf{Signal}\end{tabular}
& SR & CR
& SR & CR
& SR & CR
& SR & CR
& SR & CR
& SR & CR
& SR & CR \\

\midrule

-- & -- & --
& 0.545 & 0.323
& 0.644 & 0.437
& 0.793 & 0.670
& 1.650 & 3.018
& 0.417 & 0.320
& 0.922 & 0.859
& 0.829 & 0.938 \\

Input & -- & --
& 1.411 & 1.475
& 1.587 & 1.374
& 0.856 & 0.749
& 1.485 & 2.388
& 0.532 & 0.321
& 1.169 & 1.200
& 1.173 & 1.251 \\

\midrule

Target & Single-Market & Hidden
& 1.146 & 1.041
& 1.658 & 1.626
& 1.117 & 0.828
& 1.474 & 3.374
& 0.458 & 0.288
& 1.243 & 1.357
& 1.183 & 1.419 \\

Target & Cross-Market & Recon. Ind.
& 1.483 & 1.594
& 1.897 & 1.831
& 1.098 & 1.000
& 1.641 & 3.175
& \textbf{0.635} & \textbf{0.467}
& 1.247 & 1.333
& 1.333 & 1.567 \\

Target & Cross-Market & Hidden
& \textbf{1.520} & \textbf{1.640}
& \textbf{1.982} & \textbf{2.207}
& \textbf{1.245} & \textbf{1.146}
& \textbf{1.645} & \textbf{3.774}
& 0.568 & 0.413
& \textbf{1.418} & \textbf{1.549}
& \textbf{1.396} & \textbf{1.790} \\

\bottomrule
\end{tabular}}
\label{tab:indicator_analysis}
\end{table}

%% file: content/robuestness_efficiency.tex
\subsection{Robustness and Data Efficiency}
\label{sec:robustness_efficiency}

\paragraph{Robustness across downstream settings.}
Table~\ref{tab:robustness} compares \model{} against five best-performing methods in Table~\ref{tab:main_results} covering diverse pretraining objectives: masked reconstruction (PatchTST), contrastive learning (FASCL, TS2Vec), retrieval-augmented reconstruction (MERA), and autoregressive pretraining (Kronos-small).
These methods are evaluated across five prediction horizons under both long-only and long-short strategies, yielding ten downstream settings in total.
While MERA, PatchTST, and Kronos-small lead in specific settings, \model{} achieves the highest overall average SR (1.715) and best average rank (1.6), ranking first in five settings and second in four, making it the most consistent method across prediction horizons and portfolio strategies.

\input{table/loss_design}

\begin{figure}[ht]
  \centering
\includegraphics[width=0.48\textwidth]{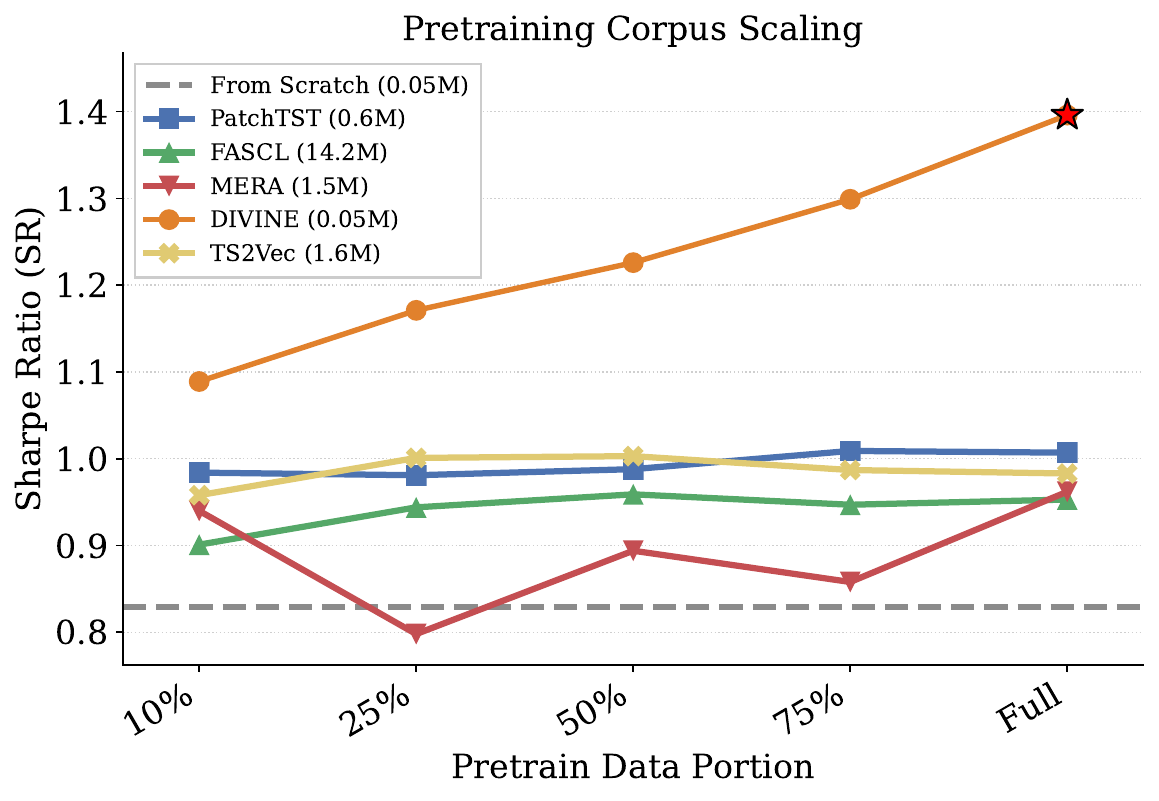}
\includegraphics[width=0.48\textwidth]{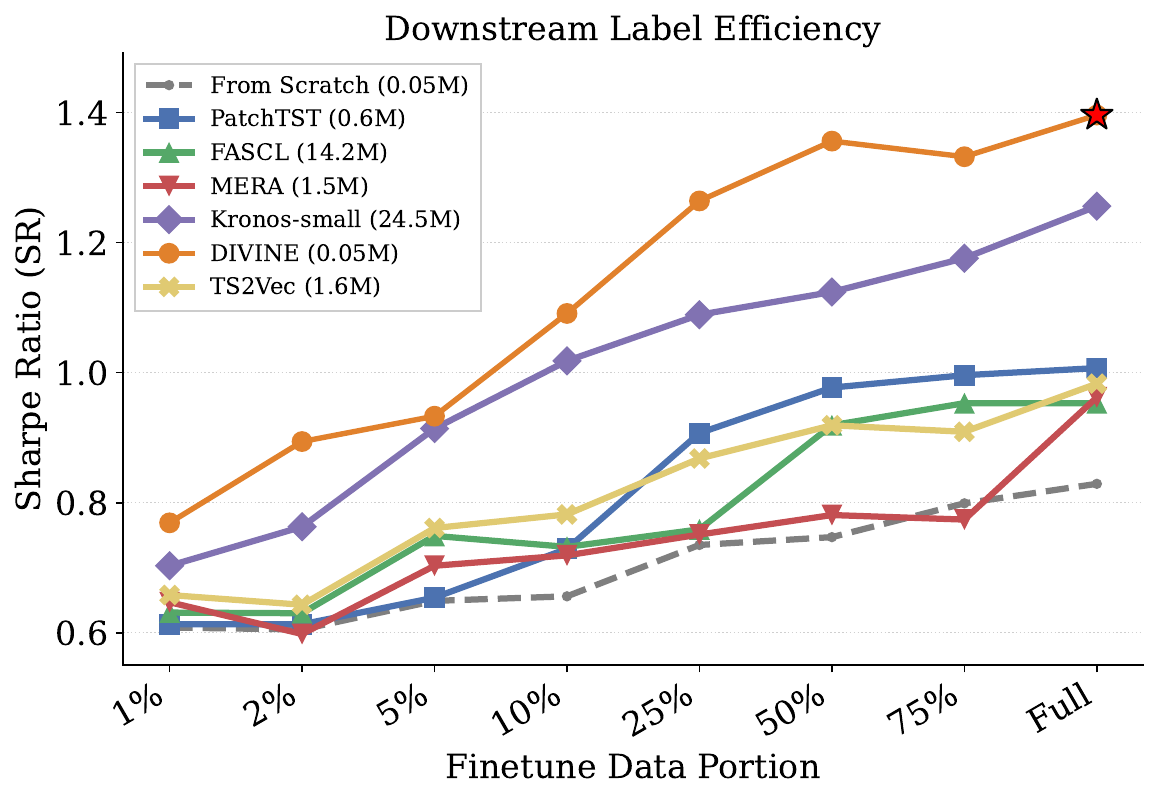}
  \caption{Pretraining-corpus scaling (left) and downstream-label efficiency (right) in average SR across six markets. Kronos-small is excluded from the left panel as an externally pretrained model.}
\label{fig:pretrain_efficiency}
\end{figure}

\paragraph{Data efficiency across pretraining and finetuning.}
Figure~\ref{fig:pretrain_efficiency} evaluates scaling with pretraining data (left) and downstream labels (right).
\model{} improves steadily with more pretraining data and, using only 10\% of the corpus, already surpasses all compared pretraining baselines trained on the full corpus.
It also leads across all downstream-label fractions and exceeds full-data training from scratch with only 5\% of the labels.
In contrast, no other pretraining method matches \model{} at any data fraction, including Kronos-mini despite its substantially larger model size and external pretraining corpus.
Similar trends hold for CR, with full results in Appendix~\ref{app:full_efficiency}.

Having established \model{}'s strong performance across horizons, strategies, and data regimes, we next examine what drives these gains:
which properties of the indicator target space matter (Section~\ref{sec:indicator_design}), and how the composition of pretraining markets shapes cross-market transfer (Section~\ref{sec:market_composition}).

%% file: table/loss_design.tex
\begin{table}[ht]
\centering

\caption{
Average Sharpe Ratio (SR) across six markets under ten downstream settings 
formed by five prediction horizons and two portfolio strategies.
Summary columns report mean SR per strategy and overall, and mean rank across all settings (lower is better).
Best and second-best results are shown in \textbf{bold} and \underline{underline}, respectively.
}

\resizebox{\textwidth}{!}{%

\begin{tabular}{l|ccccc|ccccc|cccc}

\toprule

\textbf{Portfolio Strategy}
& \multicolumn{5}{c|}{\textbf{Long-only (LO)}}
& \multicolumn{5}{c|}{\textbf{Long-Short (LS)}}
& \multicolumn{4}{c}{\textbf{Summary (Avg.)}} \\

\textbf{Prediction Horizon}
& \textbf{1D} & \textbf{3D} & \textbf{5D} & \textbf{10D} & \textbf{20D}
& \textbf{1D} & \textbf{3D} & \textbf{5D} & \textbf{10D} & \textbf{20D}
& \textbf{LO}
& \textbf{LS}
& \textbf{Overall}
& \textbf{Rank} \\

\midrule

From Scratch
& 1.345
& 0.742
& 0.829
& 0.878
& 0.802
& 1.196
& 0.508
& 0.653
& 0.666
& 0.683
& 0.919 
& 0.741 
& 0.830 
& 6.1 \\

\midrule

TS2Vec
& 2.280
& 1.171
& 0.983
& 0.766
& 0.684
& 3.026
& 1.430
& 1.062
& 0.745
& 0.559
& 1.177 
& 1.364 
& 1.271 
& 5.8
\\

PatchTST
& 2.657
& 1.190
& 1.007
& \underline{0.967}
& \textbf{0.904}
& 3.403
& 1.513
& 1.246
& 0.989
& 0.760
& 1.345
& 1.582
& 1.464
& 3.3 \\

FASCL
& 2.558
& 1.311
& 0.953
& 0.797
& 0.602
& 3.093
& 1.485
& 1.070
& 0.814
& 0.449
& 1.244 
& 1.382 
& 1.313 
& 5.5 \\

MERA
& \textbf{3.309}
& 1.376
& 0.962
& 0.900
& 0.778
& \textbf{4.233}
& 1.847
& \underline{1.462}
& \textbf{1.195}
& \textbf{0.923}
& 1.465 
& \textbf{1.932} 
& \underline{1.699} 
& \underline{2.4} \\

Kronos-small
& 3.034
& \underline{1.638}
& \underline{1.256}
& 0.892
& 0.731
& \underline{3.856}
& \underline{1.901}
& 1.337
& 0.841
& 0.545
& \underline{1.510} 
& 1.696 
& 1.603 
& 3.3 \\

\midrule

\model{}
& \underline{3.158}
& \textbf{1.688}
& \textbf{1.396}
& \textbf{1.087}
& \underline{0.850}
& 3.543
& \textbf{1.984}
& \textbf{1.524}
& \underline{1.121}
& \underline{0.801}
& \textbf{1.636}
& \underline{1.795}
& \textbf{1.715}
& \textbf{1.6} \\
\bottomrule
\end{tabular}
}
\label{tab:robustness}
\end{table}

%% file: content/analysis.tex
\section{Analysis}
\label{sec:analysis}

\subsection{What Makes a Good Indicator Set for Pretraining?}
\label{sec:indicator_design}

\begin{figure}[t]
  \centering
\includegraphics[width=0.48\textwidth]{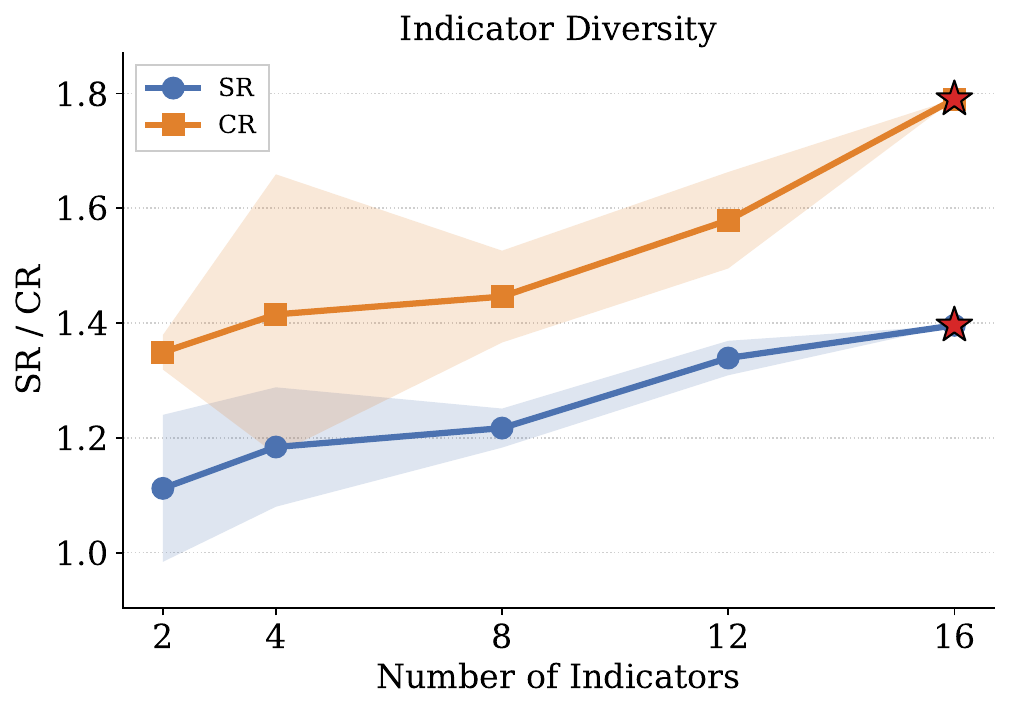}
\includegraphics[width=0.48\textwidth]{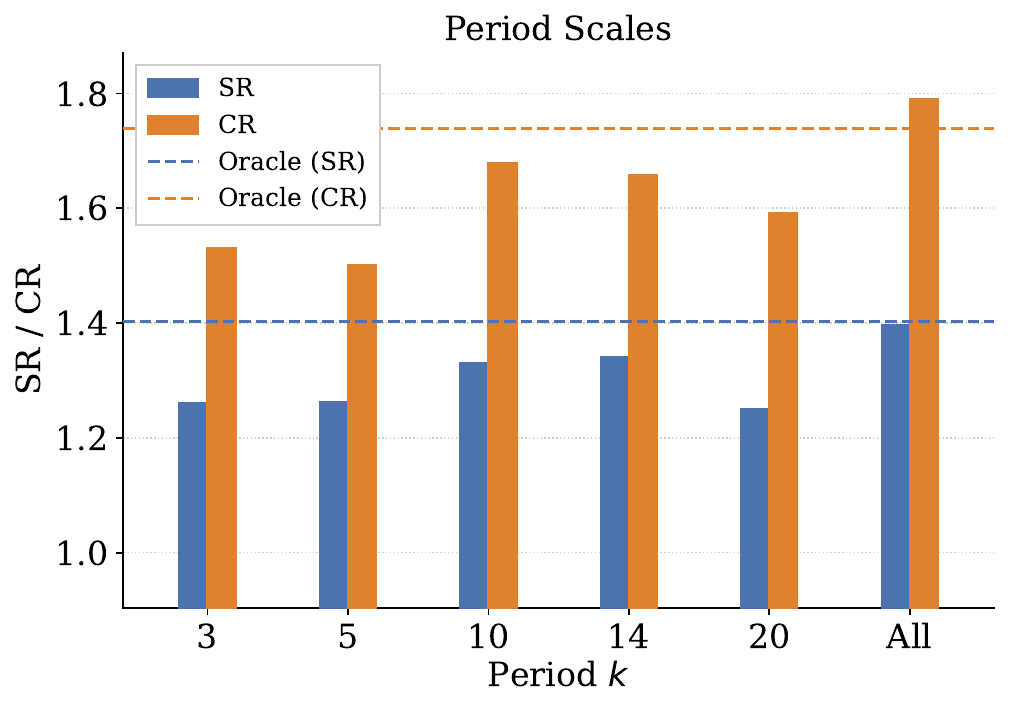}
\caption{Average SR and CR across six markets when varying the number of pretraining indicators $n$ (left; mean and std across three random draws for each $n$) and the period scale $k$ (right).}
\label{fig:good_indicator}
\end{figure}

\paragraph{Indicator diversity.}
We examine how the breadth of the supervision space affects transfer performance by randomly sampling $n\in\{2,4,8,12\}$ of the 16 indicators while retaining all period scales (Figure~\ref{fig:good_indicator}, left).
Both SR and CR improve monotonically with $n$, while standard deviation across random draws decreases, showing that broader indicator coverage yields stronger and more consistent downstream performance.

\paragraph{Temporal-scale diversity.}
Holding all 16 indicators fixed, we then vary their period scale $k$ (Figure~\ref{fig:good_indicator}, right).
No single period scale consistently outperforms others, while combining all scales performs comparably to the best single scale selected per market in hindsight (Oracle; see Appendix~\ref{app:full_indicator_design}).
This is consistent with the observation that indicators computed over nearby period  scales are inherently correlated, providing limited additional signal to the encoder beyond what a single scale already captures.

Together, these results reveal a clear design principle: \textbf{\emph{indicator-type diversity matters substantially more than period-scale diversity}}. 
Importantly, this pattern cannot be attributed simply to having more reconstruction targets: the 4-indicator setting already yields roughly 20 targets (via multiple period scales) yet still underperforms a configuration using all indicator types at a single period scale (i.e., 16 targets). The gain therefore stems from covering a broader range of distinct market dynamics, not from increasing the number of correlated targets.

\subsection{How Does Pretraining Market Composition Affect Transfer?}
\label{sec:market_composition}
\input{table/market_divesity_compact}

Building on the cross-market gains established in Section~\ref{sec:sanity_check}, Table~\ref{tab:analysis_market_compact} examines how transfer varies with the composition of the pretraining markets.
We consider both single-source pretraining and target-excluded cross-market pretraining (defined in Section~\ref{sec:problem}), where the latter provides a strict OOD setting in which the target market is never observed during pretraining.

\paragraph{Transfer to unseen markets.}
The \textit{Overall} block shows strong transfer to unseen markets.
Target-excluded pretraining outperforms single-market pretraining on all six markets, increasing average SR from 1.183 to 1.320, compared with 1.396 under joint pretraining.
Thus, 64\% of the single-market-to-joint improvement is recovered without target-market pretraining data, showing that indicator-based representations transfer effectively across markets.

\paragraph{Complementary roles of source markets.}
Transfer nevertheless varies across source--target pairs.
The \textit{One Market Only} block shows the strongest positive transfer between SP500 and TWSE300 in both directions, while the \textit{Remove One Market} block shows the largest overall drops when CSI500 or CSI300 is removed.
This distinction suggests that some markets transfer particularly well to specific targets, whereas others contribute structure that is broadly useful across markets.
CR also show a similar trend to what we observed in SR, with full results are reported in Appendix~\ref{app:full_market_composition}.

\paragraph{Toward scalable financial pretraining.}
Together with the indicator-space analysis in Section~\ref{sec:indicator_design}, these results reveal two complementary sources of transfer:
\textbf{\emph{diversity in what is reconstructed and diversity in where it is learned}}.
Diversity across indicator types broadens the supervision space, while market diversity exposes the encoder to heterogeneous market dynamics.
Combined with the data-scaling results in Section~\ref{sec:robustness_efficiency}, these findings support supervision and market diversity as key ingredients for scalable financial pretraining.

%% file: table/market_divesity_compact.tex
\definecolor{indomain}{RGB}{220,220,220}
\definecolor{transfer}{RGB}{220,228,248}
\definecolor{oodcell}{RGB}{252,234,208}

\begin{table}[t]
\small
\centering
\caption{
Average SR across six markets under different pretraining-market compositions.
\textcolor{indomain}{\rule{0.9em}{0.55em}}~Gray: in-domain training (pretrain and evaluate on the same market).
\textcolor{transfer}{\rule{0.9em}{0.55em}}~Blue: single-source transfer exceeding the target's in-domain result.
\textcolor{oodcell}{\rule{0.9em}{0.55em}}~Orange: target-excluded pretraining (target market unseen during pretraining).
Parentheses report the gap to joint pretraining ($\Delta\text{SR}=1.396-\text{Avg. SR}$).
\textbf{Bold} and \underline{underline} indicate the best value per column overall and within each block, respectively.
}
\label{tab:analysis_market_compact}

\resizebox{0.85\textwidth}{!}{%
\begin{tabular}{lccccccc}
\toprule

& \multicolumn{7}{c}{\textbf{Evaluation market}} \\
\cmidrule(lr){2-8}

\textbf{Pretraining setting} $\downarrow$ & \textbf{CSI300}
& \textbf{CSI500}
& \textbf{NI225}
& \textbf{SP500}
& \textbf{FTSE100}
& \textbf{TWSE300}
& \textbf{Avg. SR ($\Delta$SR)} \\

\midrule

\multicolumn{8}{l}{\textbf{\textit{One Market Only}}} \\

CSI300-only
& \cellcolor{indomain}\underline{1.146}
& 1.608
& 1.053
& 1.468
& 0.391
& \cellcolor{transfer}1.272
& \underline{1.158} (+.238) \\

CSI500-only
& 1.072
& \cellcolor{indomain}\underline{1.658}
& 0.954
& \cellcolor{transfer}1.516
& \cellcolor{transfer}0.463
& 1.175
& 1.140 (+.256) \\

NI225-only
& 0.633
& 1.310
& \cellcolor{indomain}\underline{1.117}
& 1.370
& \cellcolor{transfer}\underline{0.478}
& \cellcolor{transfer}1.273
& 1.047 (+.349) \\

SP500-only
& 0.543
& 1.445
& 0.950
& \cellcolor{indomain}1.474
& 0.427
& \cellcolor{transfer}\underline{1.413}
& 1.042 (+.354) \\

FTSE100-only
& 0.630
& 1.250
& 1.032
& 1.418
& \cellcolor{indomain}0.458
& \cellcolor{transfer}1.294
& 1.014 (+.382) \\

TWSE300-only
& 0.383
& 1.100
& 1.019
& \cellcolor{transfer}\underline{1.631}
& 0.402
& \cellcolor{indomain}1.243
& 0.963 (+.433) \\

\midrule

\multicolumn{8}{l}{\textbf{\textit{Remove One Market}}} \\

w/o CSI300
& \cellcolor{oodcell}1.318
& 1.875
& 1.135
& 1.491
& 0.560
& 1.344
& 1.287 (+.109) \\

w/o CSI500
& 1.054
& \cellcolor{oodcell}1.754
& 1.113
& 1.559
& \textbf{\underline{0.617}}
& \underline{1.414}
& 1.251 (+.145) \\

w/o NI225
& 1.264
& 1.969
& \cellcolor{oodcell}1.200
& 1.549
& 0.587
& 1.405
& 1.329 (+.067) \\

w/o SP500
& \underline{1.324}
& \textbf{\underline{2.129}}
& 1.204
& \cellcolor{oodcell}\textbf{\underline{1.710}}
& 0.394
& 1.403
& 1.361 (+.035) \\

w/o FTSE100
& 1.321
& 1.949
& \textbf{\underline{1.273}}
& 1.664
& \cellcolor{oodcell}0.594
& 1.412
& \underline{1.369} (+.027) \\

w/o TWSE300
& 1.245
& 1.925
& 1.188
& 1.690
& 0.522
& \cellcolor{oodcell}1.345
& 1.319 (+.077) \\

\midrule
\midrule

\multicolumn{8}{l}{\textbf{\textit{Overall}}} \\

\rowcolor{indomain}
Single Market
& 1.146
& 1.658
& 1.117
& 1.474
& 0.458
& 1.243
& 1.183 (+.213) \\

\rowcolor{oodcell}
OOD performance
& 1.318
& 1.754
& 1.200
& \underline{1.710}
& \underline{0.594}
& 1.345
& 1.320 (+.076) \\

Joint All Markets
& \textbf{\underline{1.520}}
& \underline{1.982}
& \underline{1.245}
& 1.645
& 0.568
& \textbf{\underline{1.418}}
& \textbf{\underline{1.396}} (+.000) \\

\bottomrule
\end{tabular}%
}
\end{table}

%% file: content/conclusion.tex
\section{Conclusion}
\label{sec:conclusion}
We presented \model{}, a simple cross-market pretraining framework that learns financial representations by reconstructing diverse technical indicators from raw OHLCV history. 
Across six equity markets, \model{} achieves the strongest average portfolio performance with a lightweight encoder,
outperforming existing pretraining baselines and matching or exceeding substantially larger financial foundation models, while remaining robust and data-efficient.
Our analyses further show that the gains arise from the learned representation rather than merely from exposure to technical indicators. 
More broadly, \model{} benefit from diversity in both the supervision signal and the pretraining markets: broader indicator-type coverage strengthens representation learning, while heterogeneous source markets provide reusable structure even without observing the target market during pretraining. 
Together, these findings suggest that supervision-target design and cross-market diversity offer a principled and lightweight path toward transferable financial representations.

%% file: content/appendix_indicator.tex
\section{Technical Indicator Targets}
\label{app:indicator}

Table~\ref{tab:technical_indicators} summarizes the 16 technical indicators used to construct the pretraining targets of \model{}.
All indicator values are computed using \texttt{stockstats} (version 0.6.8) \footnote{\url{https://github.com/jealous/stockstats}}, which serves as the authoritative implementation used in our experiments. 
The descriptions are intended to provide intuition about the financial quantity summarized by each target rather than to reproduce the exact numerical implementation.

Several indicators have well-established origins in the technical-analysis literature. RSI, ATR, and the directional-movement indicators DX and ADX follow Wilder~\citep{wilder1978}; MACD follows Appel~\citep{appel2005}; and CCI follows Lambert~\citep{lambert1980}. The remaining indicators are standard technical-analysis constructions summarized, for example, by Murphy~\citep{murphy1999study}.

Unless otherwise specified, we use $\mathcal{K}=\{3,5,10,14,20\}$ as the set of period configurations $k$. MACD uses its fixed $(12,26,9)$ parameterization, while RSV additionally includes $k=1$, yielding $D=77$ indicator--period targets in total.

\input{table/indicator}

\section{Datasets and Evaluation}
\label{app:data_and_eval}

\subsection{Data Acquisition and Stock Universe Construction}
\label{app:datasets}

We collect daily stock-level OHLCV data for six equity-market datasets: CSI300, CSI500, S\&P 500, Nikkei 225, FTSE 100, and Taiwan stocks. Historical data for CSI300 and CSI500 are obtained from \texttt{Qlib}, while data for S\&P 500, Nikkei 225, and FTSE 100 are retrieved from Yahoo Finance through the \texttt{yfinance} package. Taiwan stock data are obtained from \texttt{FinLab}. For each dataset, we retain the available daily open, high, low, close, and trading-volume observations within the 2008--2024 study period.

To ensure consistent temporal coverage, we apply a common coverage criterion when constructing the final stock universe for each dataset.
For the Taiwan market, we first select the 300 stocks with the highest average trading volume over the 2008--2024 study period and denote this liquidity-based candidate universe as \textnormal{TWSE300}.
We then apply the coverage filter to all datasets, retaining stocks whose trading histories span the full study period and whose missing rate across OHLCV observations is below 3\%.
All methods are trained and evaluated on the same resulting stock universe for each market to support controlled method comparisons rather than a fully point-in-time investable-universe evaluation.
Table~\ref{tab:dataset_stats} summarizes the final stock counts and temporal splits.

\input{table/dataset_stats}

\subsection{Input Features and Prediction Target}
\label{app:dataset_preprocessing}
Following previous work~\citep{sun2023alphamix,den2026integrating}, we construct an eight-dimensional input vector consisting of standardized OHLCV variables over a 20-day rolling window and three moving-average deviation features at periods $k\in\{5,10,20\}$ days. The prediction target is the forward 5-day return. Formally:
\begin{equation*}
\tilde{x}_{\text{close}}^t = \dfrac{x_{\text{close}}^t - \bar{x}_{\text{close}}^{[t-19:t]}}{\sigma_{x_{\text{close}}}^{[t-19:t]}}, \quad
\tilde{x}_{d_k}^t = \dfrac{\frac{1}{k}\sum_{i=0}^{k-1} x_{\text{close}}^{t-i}}{x_{\text{close}}^t} - 1, \quad
y_t = \dfrac{x_{\text{close}}^{t+h} - x_{\text{close}}^t}{x_{\text{close}}^t},
\end{equation*}
where $\tilde{x}_{\text{close}}^t$ denotes the normalized close price (O, H, L, and V are normalized identically), $\tilde{x}_{d_k}^t$ the moving-average deviation feature, and $y_t$ the prediction target.

\subsection{Portfolio Backtesting Protocol}
\label{app:evaluation}

We evaluate stock rankings through portfolio backtesting with top-$k$ stock selection and softmax-based portfolio weighting.
For each rebalancing day, the $k$ stocks with the highest prediction scores are selected, and a softmax over their scores determines the portfolio weights. Unless otherwise specified, $k$ is set to 5.

Because the default prediction horizon is $h=5$ trading days, each portfolio is held for $W=h=5$ trading days before rebalancing.
A single choice of the initial rebalancing day would make the resulting backtest dependent on an arbitrary calendar offset.
We therefore evaluate all $W$ possible staggered starting offsets.
Specifically, for each offset $w\in\{0,\ldots,W-1\}$, we construct a portfolio stream that starts at offset $w$ and is subsequently rebalanced every $W$ trading days.
Algorithm~\ref{alg:portfolio_returns} implements these staggered portfolio streams jointly using a circular buffer, where $\bm{Y}[w, :]$ corresponds to the daily portfolio-return series $\{r_t^{(w)}\}$ for starting offset $w$.

For each random seed $s\in\{0,\ldots,4\}$ and starting offset $w\in\{0,\ldots,4\}$, let
$\{r_t^{(s,w)}\}_{t=1}^{T_w}$ denote the resulting daily portfolio-return series.
We report the annualized Sharpe Ratio (SR) and Calmar Ratio (CR).
Let
\begin{equation}
    \bar r^{(s,w)}
    =
    \frac{1}{T_w}\sum_{t=1}^{T_w} r_t^{(s,w)},
\end{equation}
and
\begin{equation}
    \sigma^{(s,w)}
    =
    \sqrt{
        \frac{1}{T_w}
        \sum_{t=1}^{T_w}
        \left(r_t^{(s,w)}-\bar r^{(s,w)}\right)^2
    }.
\end{equation}
The Sharpe Ratio is
\begin{equation}
    \mathrm{SR}^{(s,w)}
    =
    \sqrt{252}\,
    \frac{\bar r^{(s,w)}}{\sigma^{(s,w)}}.
\end{equation}

For the Calmar Ratio, let
\begin{equation}
    C_t^{(s,w)}
    =
    \sum_{\tau=1}^{t} r_\tau^{(s,w)}
\end{equation}
denote the cumulative return and define the maximum drawdown as
\begin{equation}
    D^{(s,w)}
    =
    \max_t
    \left(
        \max_{\tau\leq t} C_\tau^{(s,w)}
        -
        C_t^{(s,w)}
    \right).
\end{equation}
The Calmar Ratio is then
\begin{equation}
    \mathrm{CR}^{(s,w)}
    =
    \frac{252\,\bar r^{(s,w)}}{D^{(s,w)}}.
\end{equation}

For each model, we evaluate five random seeds and five staggered starting offsets, yielding $5\times5=25$ seed--offset evaluations per setting:
\begin{equation}
    \overline{\mathrm{SR}}
    =
    \frac{1}{25}
    \sum_{s=0}^{4}\sum_{w=0}^{4}
    \mathrm{SR}^{(s,w)},
    \qquad
    \overline{\mathrm{CR}}
    =
    \frac{1}{25}
    \sum_{s=0}^{4}\sum_{w=0}^{4}
    \mathrm{CR}^{(s,w)}.
    \label{eq:metric_average}
\end{equation}

Unless otherwise stated, transaction costs are not included to remain consistent with the benchmark evaluation protocol.

\begin{algorithm}[H]
\caption{Portfolio Return Calculation with Staggered Rebalancing}
\label{alg:portfolio_returns}
\small
\begin{algorithmic}[1]
\Require Predictions $\bm{P} \in \mathbb{R}^{B \times N}$, next-day returns $\bm{R} \in \mathbb{R}^{B \times N}$, window length $W$
\Ensure Portfolio returns $\bm{Y} \in \mathbb{R}^{W \times B}$
\State Initialize portfolio return matrix $\bm{Y} \gets \bm{0}^{W \times B}$
\For{$d = 0$ to $B-1$} \Comment{For each day in batch}
    \State $\texttt{row\_index} \gets d \bmod W$
    \State $\texttt{time\_range} \gets \min(W, B - d)$
    \State Extract predictions $\bm{p}_d \gets \bm{P}[d, :]$ \Comment{Shape: $N$}
    \State Sort stocks: $\bm{I} \gets \text{argsort}(\bm{p}_d, \text{descending})$
    \State $\mathcal{S}_{\text{top}} \gets \bm{I}[:k]$ \Comment{Top-$k$ stock indices}
    \State $\bm{p}_{\text{top}} \gets \bm{p}_d[\mathcal{S}_{\text{top}}]$ \Comment{Top-$k$ predictions}
    \State $\bm{w} \gets \text{softmax}(\bm{p}_{\text{top}})$ \Comment{Portfolio weights}
    \State $\bm{R}_{\text{range}} \gets \bm{R}[d:d+\texttt{time\_range}, \mathcal{S}_{\text{top}}]$ \Comment{Future returns}
    \State $\bm{Y}[\texttt{row\_index}, d:d+\texttt{time\_range}] \mathrel{+}= \bm{R}_{\text{range}}^\top \bm{w}$ \Comment{Accumulate weighted returns}
\EndFor
\State \Return $\bm{Y}$
\end{algorithmic}
\end{algorithm}

\paragraph{Extension to Long--Short Portfolios.}
The same backtesting procedure can be extended to the long--short setting. The long leg is constructed exactly as in Algorithm~\ref{alg:portfolio_returns}. For the short leg, we instead select the bottom-$k$ stocks according to the same prediction scores, negate their scores before applying the softmax so that lower-ranked stocks receive larger short weights, and multiply the resulting portfolio return by $-1$. The daily long--short return is then obtained by equally averaging the returns of the long and short legs. We compute the Sharpe and Calmar ratios from this combined daily return series using the same staggered offsets and seed averaging described above.

\section{Implementation Details of \model{}}
\label{app:hydra_implementation}

\subsection{Detailed Preprocessing Algorithm (\cref{sec:targets})}
\label{app:preprocess_algo}

Technical indicators are computed once along each stock's historical series using only observations available up to each timestamp, rather than being reinitialized within each encoder input window. This preserves the standard recursive computation of indicators such as EMA and avoids window-dependent reinitialization effects.

We provide the exact preprocessing procedure used to construct the normalized indicator targets in Algorithm~\ref{alg:preprocess}. All statistics used for clipping and standardization are estimated from the training split only and are kept fixed for validation and testing.
The procedure consists of three stages: sample-level scaling of OHLCV and scale-dependent indicator targets, per-market clipping and standardization, and pooled re-standardization for joint cross-market pretraining.
It is worth noting that scale-free indicators are not affected by the sample-level price normalization, while price-dependent targets are divided by the corresponding within-window maximum price. Table~\ref{tab:technical_indicators} lists the classification of each target.

\begin{algorithm}[H]
\caption{Target Normalization, PyTorch-style}
\small
\label{alg:preprocess}
\begin{lstlisting}[style=pytorch, basicstyle=\ttfamily\footnotesize]
# X: (N, L, 5) OHLCV input windows for market m
# Z: (N, D) raw indicator targets for market m
# price_idx: indices of price-based target dimensions
# scale_free_idx: indices of scale-free target dimensions
# Training split only; statistics are frozen at inference.

def per_sample_max_norm(X, Z):
    p_max = X[:, :, :4].max(dim=1).values.max(dim=1).values  # (N,)
    v_max = X[:, :, 4].max(dim=1).values                      # (N,)
    X[:, :, :4] = X[:, :, :4] / p_max[:, None, None]
    X[:, :, 4]  = X[:, :, 4]  / v_max[:, None]
    Z[:, price_idx] = Z[:, price_idx] / p_max[:, None]
    return X, Z  # scale-free dims untouched

def per_market_clip_and_znorm(Z):
    # statistics computed over all training samples (N,) per dim
    med = Z.median(dim=0).values                # (D,)
    iqr = Z.quantile(0.75, dim=0) - Z.quantile(0.25, dim=0)  # (D,)
    Z = Z.clip(med - 10 * iqr, med + 10 * iqr)  # IQR clipping
    mu, sigma = Z.mean(dim=0), Z.std(dim=0)      # (D,)
    Z = (Z - mu) / (sigma + eps)                  # z-norm
    return Z, mu, sigma

# --- Stage 1: per-sample max-normalization ---
X, Z = per_sample_max_norm(X, Z)

# --- Stage 2: per-market IQR clipping + z-normalization ---
Z, mu_m, sigma_m = per_market_clip_and_znorm(Z)

# --- Stage 3 (cross-market only): pooled re-standardization ---
# After concatenating all markets:
# Z_all: (N_total, D), padded to common cross-section width
mu_pool, sigma_pool = Z_all.mean(dim=0), Z_all.std(dim=0)
Z_all = (Z_all - mu_pool) / (sigma_pool + eps)
\end{lstlisting}
\end{algorithm}

\subsection{Architecture and Training Configuration}
\label{app:hydra_hyperparams}

Table~\ref{tab:hydra_training} summarizes the optimization and transfer settings used for \model{}, while Table~\ref{tab:hydra_architecture} reports the architecture-specific configurations of the GRU, Transformer, and Mamba encoders. Unless otherwise stated, all encoder variants use the same indicator reconstruction objective, preprocessing pipeline, and downstream ranking protocol.

During pretraining, the encoder and a randomly initialized linear reconstruction head are optimized jointly to reconstruct the 77 normalized indicator targets from a 120-day OHLCV context. 
Each training sample is one trading day's full stock cross-section, and a batch of 256 days may draw from different markets under cross-market pretraining. 
Because stock universes differ in size across markets, cross-sections are zero-padded to a common width and the padded entries are excluded from the reconstruction loss. Batches are accumulated over 16 steps of 16 days each.
For all backbones, the final sequence representation is mapped to the same 77-dimensional target space by this linear head. After pretraining, the reconstruction head is discarded and replaced by a two-layer market-specific MLP with hidden dimension 64 for downstream stock ranking.

\model{} is transferred with a frozen encoder by default. Freezing isolates the contribution of the pretrained representation, so that differences across pretraining settings in Section~\ref{sec:sanity_check} reflect the pretraining objective rather than downstream adaptation.
The Transformer encoder is the sole exception, arising from two interacting factors:
\textbf{(1)} the gap between the 120-day pretraining context length and 20-day downstream windows, and 
\textbf{(2)} the use of sinusoidal positional encoding. 
Because the representation is taken from the last token, the positional encoding assigns the downstream representation a position index that differs from the one it occupies during pretraining. 
We offset the downstream positions so that the 20-day window occupies the final positions of the pretraining context (shifting the position index from 0--19 to 100--119), without which transfer degrades severely.
Even with this correction, frozen transfer remains weaker than full fine-tuning for the Transformer. We therefore fully fine-tune the Transformer encoder, allowing its positional representations to adapt to the shorter downstream window. Results under both transfer protocols are reported in Appendix~\ref{app:main_with_freeze}.
In constrast, the recurrent and state-space encoders carry no absolute positional encoding and are therefore unaffected.
Adapting the framework to architectures with absolute positional encodings, for instance through relative or bias-based position representations, remains an open direction.

For both pretraining and downstream transfer, we use AdamW with a fixed learning rate of $10^{-3}$ and no weight decay or learning-rate scheduling. Both stages are trained for 100 epochs with an effective batch size of 256, without early stopping. Gradients are accumulated over 16 steps and clipped to a maximum norm of 1.0. Final performance is averaged over five random seeds. All experiments are implemented in PyTorch and run in full precision on NVIDIA RTX 3090 GPUs.

\input{table/hydra_hyperparam}

\section{Implementation Details of Baselines}
\label{app:baseline_implementation}

\subsection{Generic and Financial Pretraining Methods}

\paragraph{Transfer Protocol.}
We follow each method's original transfer protocol whenever it can be applied directly to cross-sectional ranking, and otherwise use full fine-tuning to give the encoder the maximum capacity to adapt. 
In practice this yields full fine-tuning for all pretraining baselines, though for two different reasons. PatchTST, MERA, and CI-STHPAN already fine-tune the full encoder in their original protocols, so we follow them directly. 
The remaining methods depart from the ranking objective and prediction head used here, and therefore fall under the second case: TS2Vec and TimesURL evaluate frozen representations with linear or kernel-based heads, FASCL retrieves stocks by embedding similarity, and SSPT selects its freezing strategy by hyperparameter search. 
All baselines are transferred with a shared MLP head identical to that used in \model{}. 
For completeness, we additionally evaluate all pretraining methods under both frozen-encoder transfer and full fine-tuning in Appendix~\ref{app:main_with_freeze}.

\paragraph{Implementation Details and Hyperparameter Settings.}
Table~\ref{tab:pretrain} summarizes the pretraining configuration of all baselines. We follow each method's released configuration where one is specified, and otherwise set the remaining hyperparameters to values consistent with our evaluation setting. 
Four cases require further explanation. FASCL has no released implementation, so we reproduce it from the paper using the reported 
hyperparameters.
For TimesURL we adopt the effective command-line default learning rate of $10^{-4}$, as the repository specifies inconsistent values elsewhere. 
For PatchTST we disable RevIN, because our input windows are already temporally $z$-scored and we require identical preprocessing across encoders. 
Finally, PatchTST and CI-STHPAN both specify long self-supervised context length ($L=512$) and 
correspondingly large patch sizes. Since all methods share a 20-day downstream input window in our evaluation, retaining these settings would introduce a substantial mismatch between the pretraining and transfer stages, and would leave only one or two patches per downstream window. We therefore shorten the pretraining context length and use non-overlapping patches of $P=S=4$.

\begin{table}[ht]
\centering
\caption{Pretraining configuration of the compared pretraining baselines. Batch size follows each method's own sampling unit, where one day denotes the full stock cross-section of a single trading day.}
\label{tab:pretrain_config}
\resizebox{0.8\textwidth}{!}{
\begin{tabular}{lcccccc}
\toprule
\textbf{Method} & \textbf{Context} & \textbf{Emb.\ dim} & \textbf{Batch size} & \textbf{Learning rate} & \textbf{Training budget} & \textbf{Code} \\
\midrule
FASCL     & 64 & 384   & 1 day           & $10^{-3}\!\to\!10^{-6}$ & 50 epochs &  --- \\
TS2Vec    & 64 & 320   & 16 windows      & $10^{-3}$ & 600 iters & \href{https://github.com/zhihanyue/ts2vec}{GitHub} \\
TimesURL  & 64 & 320   & 8 windows       & $10^{-4}$ & 600 iters & \href{https://github.com/Alrash/TimesURL}{GitHub} \\
SSPT      & 32 & 128   & 1024 windows & $10^{-4}$ & 100 epochs & \href{https://github.com/astudentuser/Pre-training-Time-Series-Models-with-Stock-Data-Customization}{GitHub} \\
PatchTST  & 64 & 1024 & 64 windows    & LR range test & 100 epochs & \href{https://github.com/yuqinie98/PatchTST}{GitHub} \\
MERA      & 20 & 128   & 2 days          & $10^{-4}$ & 500 epochs & \href{https://github.com/chenchen1104/MERA}{GitHub} \\
CI-STHPAN & 20 & 128   & 1 day           & LR range test & 100 epochs & \href{https://github.com/Harryx2019/CI-STHPAN}{GitHub} \\
\bottomrule
\end{tabular}
}
\label{tab:pretrain}
\end{table}

\subsection{Financial Foundation Model}
\label{app:foundation_models}

\paragraph{Pretraining Exposure.}
The released financial foundation models are pretrained on substantially larger external corpora than the six-market corpus used by \model{}.
Kronos~\citep{shi2025kronos} is pretrained on more than 12 billion K-line records from 45 global exchanges through June 2024, resulting in temporal overlap with our 2021--2024 evaluation period.
FinCast~\citep{zhu2025fincast} is pretrained on 2.4 million time series spanning financial and non-financial domains; however, its released documentation does not provide sufficient temporal coverage information to determine whether a similar overlap exists.

\paragraph{Adaptation and Transfer Protocol.}
For both models, we use the preprocessing pipelines specified in their original publications~\footnote{\url{https://github.com/shiyu-coder/Kronos}; \url{https://github.com/vincent05r/FinCast-fts}} rather than applying our shared window normalization. 

\textit{Kronos.}
We evaluate Kronos-mini, Kronos-small, and Kronos-base using their released pretrained checkpoints.
For each downstream market, we follow the released target-market adaptation procedure, selecting the checkpoint with the lowest validation loss on the corresponding training split.
We then transfer the adapted sequence representation to the downstream 5-day cross-sectional ranking objective.
We select the \textit{timestamp-free representation transfer} configuration as our primary evaluation setting, as it is most aligned with the other baselines in our benchmark: representation transfer mirrors how other pretrained encoders are evaluated (rather than relying on Kronos's native autoregressive interface), and omitting timestamp conditioning removes an informational advantage not available to other methods.
For completeness, we additionally evaluate timestamp-conditioned representation transfer and direct autoregressive forecasting; results under all three configurations are reported in Table~\ref{tab:kronos_protocol}.

\textit{FinCast.}
No target-domain adaptation pipeline is publicly provided; as the repository explicitly supports zero-shot transfer, we initialize directly from the released pretrained checkpoint.
Because FinCast is a univariate model, encoding only a subset of input channels would place it at an informational disadvantage relative to other baselines that consume the full OHLCV bar.
To fairly leverage all available information without retraining the model, we encode each input channel independently using the pretrained backbone and concatenate the resulting representations before the downstream MLP, using the representation at the final timestep as the model output.
We evaluate three channel configurations (C, CV, and OHLCV) to ablate the contribution of each information source; results are reported in Table~\ref{tab:fincast_variants}, with the full OHLCV configuration used in Table~\ref{tab:main_results} as the most informationally complete and fairly comparable setting.

\paragraph{Full Results.}
Tables~\ref{tab:kronos_protocol} and~\ref{tab:fincast_variants} report full results under all evaluated configurations for Kronos and FinCast, respectively.
For Kronos, representation transfer substantially outperforms direct autoregressive inference, and removing timestamp conditioning—which aligns Kronos with the information available to other baselines—consistently improves performance across all three model scales.
For FinCast, incorporating additional channels improves transfer performance, with the full OHLCV variant achieving the best average SR and CR; this is also the most informationally complete configuration and therefore the most comparable to other methods.
Notably, the configuration selected for Table~\ref{tab:main_results} in each case is both the most methodologically aligned with other baselines \textit{and} the strongest-performing setting, suggesting our protocol choices do not inadvertently favor or penalize these models.

\input{table/full_foundation_model_results}

\input{content/appendix_detailed_results}

\input{content/appendix_extra_analysis}

%% file: table/indicator.tex
\begin{table}[H]
\centering
\caption{
Technical indicators used as pretraining targets.
Descriptions summarize the financial quantity captured by each indicator;
all numerical values follow the \texttt{stockstats} v0.6.8 implementation.
The Scale-free column indicates whether a target is invariant to the overall price level (Y) and therefore unaffected by the sample-level normalization in Appendix~\ref{app:preprocess_algo}.
}
\label{tab:technical_indicators}
\scriptsize
\setlength{\tabcolsep}{3pt}
\renewcommand{\arraystretch}{1.15}

\begin{tabularx}{\linewidth}{
>{\raggedright\arraybackslash}p{0.32\linewidth}
>{\raggedright\arraybackslash}X
>{\centering\arraybackslash}p{0.1\linewidth}
>{\centering\arraybackslash}p{0.08\linewidth}}
\toprule
\textbf{Indicator} & \textbf{Description} & \textbf{Period} $k$ & \textbf{Scale-free} \\
\midrule

Simple Moving Average (SMA) &
Mean closing price over the recent $k$ periods. &
$k \in \mathcal{K}$ & N \\

Exponential Moving Average (EMA) &
Exponentially weighted average of recent closing prices, assigning greater
weight to more recent observations. &
$k \in \mathcal{K}$ & N \\

Moving Average Convergence Divergence (MACD) &
Difference between short- and long-horizon exponential moving averages,
summarizing changes in trend and momentum. &
$(12,26,9)$ & N \\

True Range (TR) &
Recent trading-range magnitude accounting for both intraday high--low
variation and gaps from the previous close. &
$k \in \mathcal{K}$ & N \\

Average True Range (ATR) &
Smoothed measure of recent true range, commonly used to summarize
price volatility. &
$k \in \mathcal{K}$ & N \\

Rate of Change (ROC) &
Relative change in closing price over the preceding $k$ periods. &
$k \in \mathcal{K}$ & Y \\

Relative Strength Index (RSI) &
Relative magnitude of recent upward and downward price movements. &
$k \in \mathcal{K}$ & Y \\

Triple Exponential Average (TRIX) &
Rate of change of a triple-smoothed exponential moving average, capturing
smoothed price momentum. &
$k \in \mathcal{K}$ & Y \\

TRIX Moving Average (TRMA) &
Moving average of TRIX, providing a smoother view of TRIX dynamics. &
$k \in \mathcal{K}$ & Y \\

Raw Stochastic Value (RSV) &
Position of the current close within the recent high--low trading range. &
$k \in \mathcal{K} \cup \{1\}$ & Y \\

Stochastic \%K &
Smoothed stochastic value derived from RSV. &
$k \in \mathcal{K}$ & Y \\

Stochastic \%D &
Further smoothed version of Stochastic \%K. &
$k \in \mathcal{K}$ & Y \\

Directional Index (DX) &
Strength of directional price movement over the recent $k$ periods. &
$k \in \mathcal{K}$ & Y \\

Average Directional Index (ADX) &
Smoothed measure of directional-movement strength. &
$k \in \mathcal{K}$ & Y \\

Commodity Channel Index (CCI) &
Deviation of typical price from its recent moving average relative to its
typical dispersion. &
$k \in \mathcal{K}$ & Y \\

Money Flow Index (MFI) &
Price--volume momentum measure combining typical price changes with
trading volume. &
$k \in \mathcal{K}$ & Y \\

\bottomrule
\end{tabularx}
\end{table}

%% file: table/dataset_stats.tex
\begin{table}[H]
    \centering
    \caption{Dataset statistics after stock-universe filtering.}
    \scriptsize
    \resizebox{0.7\textwidth}{!}{%
    \begin{tabular}{lccccc}
        \toprule
        \textbf{Market} & \textbf{Region} & \textbf{\# Stocks}
        & \textbf{Train} & \textbf{Validation} & \textbf{Test} \\
        \midrule
        CSI300  & CN & 295 & \multirow{6}{*}{2008--2019} & \multirow{6}{*}{2020} & \multirow{6}{*}{2021--2024} \\
        CSI500  & CN & 514 \\
        NI225   & JP & 209 \\
        SP500   & US & 525 \\
        FTSE100 & UK & 81  \\
        TWSE300 & TW & 205 \\
        \bottomrule
    \end{tabular}
    }
    \label{tab:dataset_stats}
\end{table}

%% file: table/hydra_hyperparam.tex
\begin{table}[ht]
\centering
\caption{
Training and transfer configuration of \model{}. Settings are shared across encoder backbones unless otherwise specified.
}
\label{tab:hydra_training}
\resizebox{0.7\textwidth}{!}{
\begin{tabular}{llc}
\toprule
\textbf{Stage} & \textbf{Configuration} & \textbf{Value} \\
\midrule

\multirow{4}{*}{Pretraining}
& Input features & OHLCV \\
& Context length & 120 days \\
& Reconstruction targets & 77 \\
& Reconstruction head & Linear \\
& Loss & Mean squared error \\
\midrule

\multirow{6}{*}{Downstream transfer}
& Input features & OHLCV \\
& Input window & 20 days \\
& Prediction horizon & 5 days \\
& Prediction head & 2-layer MLP \\
& Encoder update & Frozen by default \\
& Ranking objective & Differentiable Spearman loss \\
\midrule

\multirow{10}{*}{Shared Settings}
& Optimizer & AdamW \\
& Learning rate & $10^{-3}$ \\
& Weight decay & 0.0 \\
& Effective batch size & 256 days \\
& Training epochs & 100 \\
& Gradient accumulation & 16 steps (16 days each) \\
& Gradient clipping & 1.0 \\
& LR scheduler & None \\
& Early stopping & None \\
& Random seeds & $\{0,1,2,3,4\}$ \\

\bottomrule
\end{tabular}
}
\end{table}

\begin{table}[ht]
\centering
\caption{
Architecture-specific configurations of the GRU, Transformer, and Mamba
encoders used in \model{}.
}
\label{tab:hydra_architecture}
\resizebox{0.7\textwidth}{!}{
\begin{tabular}{lccc}
\toprule
\textbf{Configuration}
& \textbf{GRU}
& \textbf{Transformer}
& \textbf{Mamba} \\
\midrule

Input dimension
& 5 & 5 & 5 \\

Hidden dimension
& 64 & 64 & 64 \\

Number of layers
& 2 & 2 & 2 \\

Attention heads
& -- & 8 & -- \\

Attention mask
& -- & Causal & -- \\

Positional encoding
& -- & Sinusoidal & -- \\

FFN dimension
& -- & 256 & -- \\

State dimension
& -- & -- & 16 \\

Convolution width
& -- & -- & 4 \\

Expansion factor
& -- & -- & 2 \\

Dropout
& 0.0 & 0.1 & 0.1 \\

Representation pooling
& Last hidden state
& Last token
& Last hidden state \\

Downstream MLP hidden dim.
& 64 & 64 & 64 \\

Downstream encoder update
& Frozen
& Full fine-tuning
& Frozen \\

\midrule

Encoder parameters (M)
& 0.05 & 0.10 & 0.07 \\

\bottomrule
\end{tabular}
}
\end{table}

%% file: table/full_foundation_model_results.tex
\begin{table*}[ht]
\centering
\caption{
Kronos results across representation-transfer and native autoregressive configurations. ``Transfer'' denotes representation transfer to the downstream ranking task, ``AR'' denotes native autoregressive inference, and ``w/ stamp'' / ``w/o stamp'' indicate whether timestamp conditioning is used.}
\label{tab:kronos_protocol}
\resizebox{\textwidth}{!}{
\begin{tabular}{ll|cc|cc|cc|cc|cc|cc|cc}
\toprule
\textbf{Scale} &
\textbf{Protocol} &
\multicolumn{2}{c|}{\textbf{CSI300}} &
\multicolumn{2}{c|}{\textbf{CSI500}} &
\multicolumn{2}{c|}{\textbf{NI225}} &
\multicolumn{2}{c|}{\textbf{SP500}} &
\multicolumn{2}{c|}{\textbf{FTSE100}} &
\multicolumn{2}{c|}{\textbf{TWSE300}} &
\multicolumn{2}{c}{\textbf{Avg.}} \\
& & SR & CR & SR & CR & SR & CR & SR & CR & SR & CR & SR & CR & SR & CR \\
\midrule

\multirow{4}{*}{Mini}
& Transfer, w/o stamp
& 1.260 & 1.344
& 1.637 & 1.822
& 0.762 & 0.592
& 1.266 & 1.644
& 0.522 & 0.365
& 1.214 & 1.085
& \textbf{1.110} & \textbf{1.142} \\

& Transfer, w/ stamp
& 0.969 & 0.843
& 1.462 & 1.398
& 0.788 & 0.668
& 0.925 & 0.942
& 0.445 & 0.294
& 1.094 & 0.883
& 0.947 & 0.838 \\

& AR, w/o stamp
& -0.043 & 0.003
& -0.572 & -0.162
& 0.728 & 0.627
& 1.340 & 2.136
& 0.568 & 0.413
& 0.626 & 0.418
& 0.441 & 0.573 \\

& AR, w/ stamp
& 0.140 & 0.085
& -0.404 & -0.123
& 0.763 & 0.660
& 1.391 & 2.453
& 0.419 & 0.327
& 0.651 & 0.412
& 0.493 & 0.636 \\

\midrule

\multirow{4}{*}{Small}
& Transfer, w/o stamp
& 1.459 & 1.516
& 1.916 & 2.229
& 0.903 & 0.783
& 1.523 & 2.329
& 0.524 & 0.378
& 1.209 & 1.186
& \textbf{1.256} & \textbf{1.404} \\

& Transfer, w/ stamp
& 1.171 & 1.160
& 1.598 & 1.565
& 0.794 & 0.651
& 1.224 & 1.622
& 0.597 & 0.423
& 1.118 & 0.972
& 1.084 & 1.066 \\

& AR, w/o stamp
& 0.152 & 0.125
& -0.275 & -0.095
& 0.575 & 0.485
& 1.268 & 1.963
& 0.494 & 0.371
& 0.592 & 0.384
& 0.468 & 0.539 \\

& AR, w/ stamp
& 0.196 & 0.143
& -0.115 & -0.030
& 0.573 & 0.482
& 1.255 & 1.805
& 0.451 & 0.339
& 0.595 & 0.370
& 0.492 & 0.518 \\

\midrule

\multirow{4}{*}{Base}
& Transfer, w/o stamp
& 1.326 & 1.305
& 1.803 & 1.979
& 0.895 & 0.714
& 1.387 & 2.040
& 0.359 & 0.252
& 1.154 & 1.076
& \textbf{1.154} & \textbf{1.228} \\

& Transfer, w/ stamp
& 1.034 & 0.856
& 1.506 & 1.328
& 0.797 & 0.713
& 1.045 & 1.085
& 0.542 & 0.370
& 1.244 & 1.117
& 1.028 & 0.912 \\

& AR, w/o stamp
& 0.011 & 0.052
& -0.684 & -0.161
& 0.666 & 0.554
& 1.345 & 2.063
& 0.486 & 0.342
& 0.605 & 0.368
& 0.405 & 0.537 \\

& AR, w/ stamp
& 0.245 & 0.181
& -0.074 & -0.007
& 0.596 & 0.504
& 1.329 & 1.846
& 0.578 & 0.445
& 0.641 & 0.429
& 0.553 & 0.566 \\

\bottomrule
\end{tabular}
}
\end{table*}

\begin{table*}[h]
\centering
\caption{
FinCast representation-transfer results across input-channel configurations. C, CV, and OHLCV denote close-only, close-volume, and full OHLCV inputs, respectively.}
\label{tab:fincast_variants}
\resizebox{\textwidth}{!}{
\begin{tabular}{l|cc|cc|cc|cc|cc|cc|cc}
\toprule
\textbf{Channels} &
\multicolumn{2}{c|}{\textbf{CSI300}} &
\multicolumn{2}{c|}{\textbf{CSI500}} &
\multicolumn{2}{c|}{\textbf{NI225}} &
\multicolumn{2}{c|}{\textbf{SP500}} &
\multicolumn{2}{c|}{\textbf{FTSE100}} &
\multicolumn{2}{c|}{\textbf{TWSE300}} &
\multicolumn{2}{c}{\textbf{Avg.}} \\
& SR & CR & SR & CR & SR & CR & SR & CR & SR & CR & SR & CR & SR & CR \\
\midrule

C
& 0.748 & 0.584
& 0.627 & 0.441
& 0.689 & 0.550
& 0.633 & 0.506
& 0.494 & 0.307
& 0.797 & 0.590
& 0.665 & 0.496 \\

CV
& 0.907 & 0.759
& 0.729 & 0.566
& 0.687 & 0.533
& 0.755 & 0.613
& 0.493 & 0.339
& 0.851 & 0.594
& 0.737 & 0.567 \\

OHLCV
& 0.780 & 0.668
& 1.036 & 0.819
& 0.649 & 0.507
& 0.699 & 0.560
& 0.434 & 0.274
& 0.924 & 0.673
& \textbf{0.754} & \textbf{0.583} \\

\bottomrule
\end{tabular}
}
\end{table*}

%% file: content/appendix_detailed_results.tex
\section{Detailed Results of Main Paper}
\label{app:detailed_results}

\subsection{Per-Indicator Reconstruction Quality (from Section~\ref{sec:objective})}
\label{app:reconstruction}
\begin{figure}[ht]
  \centering
  \includegraphics[width=\textwidth]{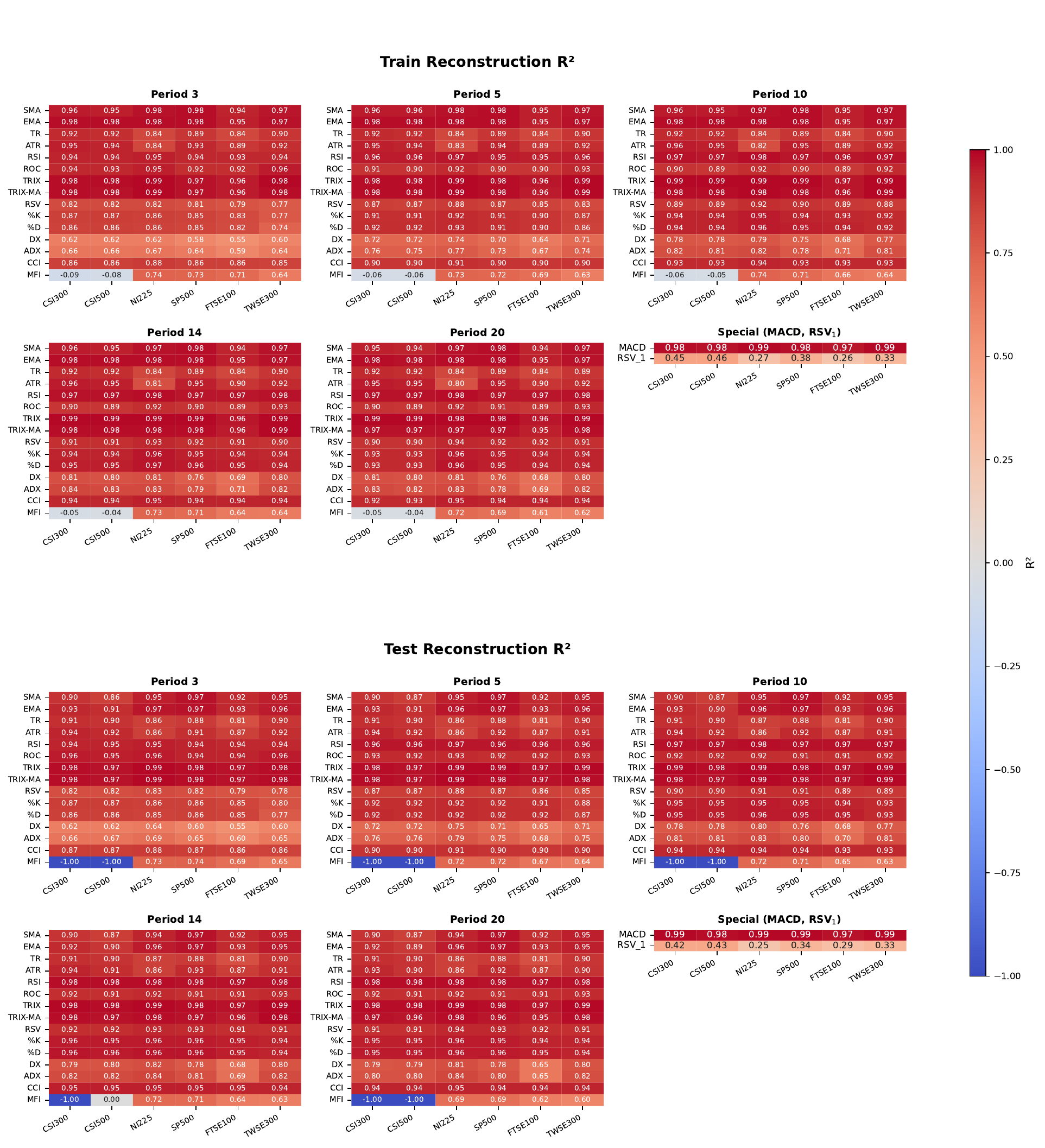}
  \caption{Per-indicator reconstruction $R^2$ on train (upper block) and
  test (lower block) splits. Rows correspond to the five period
  parameters $k \in \{3, 5, 10, 14, 20\}$; the bottom row contains the
  two special cases (MACD with fixed parameterization, RSV additionally
  computed at $k=1$). Each cell shows $R^2$ for one indicator--market
  combination. $R^2$ is clipped to $[-1, 1]$ for visualization.}
  \label{fig:reconstruction_r2}
\end{figure}

Figure~\ref{fig:reconstruction_r2} reports per-indicator reconstruction $R^2$ on both the training and test splits, decomposed by period parameter and market.
Overall, the encoder reliably reconstructs the pretraining targets: across the 77 indicator--period combinations and six markets, average test $R^2$ reaches 0.847, with median train--test gap of 0.008 and all gaps below 0.05, confirming that pretraining converges without severe overfitting.

\paragraph{MFI failure on CSI markets.}
One systematic exception stands out: MFI reconstruction fails on CSI300 and CSI500 across all period parameters (test $R^2 < 0$), while remaining comparable to other indicators on the four non-Chinese markets. MFI is computed as the ratio of positive money flow to total money flow over the window, and in the CSI markets this ratio concentrates tightly around 0.5, leaving almost no variation for the encoder to reconstruct. The remaining 76 targets are unaffected, with reconstruction quality comparable to the other markets.

Notably, despite this failure, downstream ranking performance on CSI300 and CSI500 remains the strongest among all six markets (Table~\ref{tab:main_results}). 
This dissociation between per-indicator reconstruction quality and downstream utility reinforces the observation from Section~\ref{sec:sanity_check}: the hidden state encodes ranking-relevant structure that is not tightly coupled to the reconstruction accuracy of any individual indicator.

\subsection{Additional results of Data Efficiency (Section~\ref{sec:robustness_efficiency})}
\label{app:full_efficiency}
Table~\ref{tab:pretrain_efficiency} and~\ref{tab:finetune_efficiency} report the average Calmar Ratio (CR) across six markets under varying pretraining-data and downstream-label fractions, complementing the Sharpe Ratio results reported in Section~\ref{sec:robustness_efficiency}.
The CR results exhibit trends consistent with SR: \model{} improves as more pretraining data become available and maintains strong performance as the amount of downstream supervision is reduced.
\input{table/data_efficiency}

\subsection{Additional Results on Period Diversity of Indicators (Section~\ref{sec:indicator_design})}
\label{app:full_indicator_design}
Table~\ref{tab:indicators_scaling} reports the per-market SR and CR for each single-period configuration, together with the full multi-period setting.
To construct the per-market oracle, we retrospectively select, for each market and each metric, the best-performing single period among $k \in \{3,5,10,14,20\}$, and then average the selected values across the six markets.
The oracle therefore represents an optimistic upper bound achievable by choosing the period separately for each market with hindsight, rather than a deployable model-selection procedure.

The full multi-period configuration performs comparably to this oracle, supporting the observation in Section~\ref{sec:indicator_design} that combining multiple period scales primarily improves robustness to period choice rather than providing a large additional performance gain.
\input{table/indicator_scaling}

\subsection{Additional Results on Market Composition (\cref{sec:market_composition})}
\label{app:full_market_composition}
\Cref{tab:analysis_market} reports the corresponding Calmar Ratio (CR) results under different pretraining-market compositions. The results exhibit a pattern consistent with the Sharpe Ratio (SR) analysis in \cref{sec:market_composition}.
\input{table/market_diversity}

%% file: table/data_efficiency.tex
\begin{table}[ht]
\centering
\caption{Average CR across market by pretrain efficiency}
\resizebox{0.6\textwidth}{!}{%
\begin{tabular}{lccccc}
\toprule
\textbf{Pretrain Data Portion} 
  & \multicolumn{1}{c}{\textbf{10\%}}
  & \multicolumn{1}{c}{\textbf{25\%}} 
  & \multicolumn{1}{c}{\textbf{50\%}} 
  & \multicolumn{1}{c}{\textbf{75\%}} 
  & \multicolumn{1}{c}{\textbf{Full}} \\
\midrule
From Scratch & -- & -- & -- & -- & 0.938 \\
\midrule
TS2Vec & 0.916 & 0.953 & 0.951 & 0.917 & 0.923\\
PatchTST & 1.173 & 1.206 & 1.210 & 1.229 & 1.203 \\
FASCL & 0.863 & 0.908 & 0.947 & 0.927 & 0.933 \\
MERA & 0.988 & 0.986 & 0.842 & 0.955 & 0.833  \\
\midrule
\model{} & 1.257 & 1.535 & 1.619 & 1.605 & \textbf{1.790} \\
\bottomrule
\label{tab:pretrain_efficiency}
\end{tabular}
}
\end{table}

\begin{table}[ht]
\centering
\caption{Average CR across market by finetune efficiency}
\resizebox{0.8\textwidth}{!}{%
\begin{tabular}{lcccccccc}
\toprule
\textbf{Finetune Data Portion} 
  & \multicolumn{1}{c}{\textbf{1\%}}
  & \multicolumn{1}{c}{\textbf{2\%}}
  & \multicolumn{1}{c}{\textbf{5\%}}
  & \multicolumn{1}{c}{\textbf{10\%}}
  & \multicolumn{1}{c}{\textbf{25\%}} 
  & \multicolumn{1}{c}{\textbf{50\%}} 
  & \multicolumn{1}{c}{\textbf{75\%}} 
  & \multicolumn{1}{c}{\textbf{Full}} \\
\midrule
From Scratch & 0.686 & 0.670 & 0.679 & 0.735 & 0.818 & 0.826 & 0.900 & 0.938 \\
\midrule
TS2Vec & 0.573 & 0.535 & 0.654 & 0.670 & 0.785 & 0.853 & 0.832 & 0.923 \\
PatchTST & 0.560 & 0.512 & 0.619 & 0.642 & 1.026 & 1.155 & 1.180 & 1.203 \\
FASCL & 0.511 & 0.507 & 0.629 & 0.613 & 0.684 & 0.849 & 0.923 & 0.933\\
MERA & 0.632 & 0.537 & 0.712 & 0.782 & 0.822 & 0.870 & 0.779 & 0.833 \\
Kronos-small& 0.627 & 0.662 & 0.773 & 0.963 & 1.107 & 1.158 & 1.229 & 1.404 \\
\midrule
\model{} & 0.800 & 0.910 & 0.987 & 1.233 & 1.466 & 1.652 & 1.687 & \textbf{1.790} \\
\bottomrule
\label{tab:finetune_efficiency}
\end{tabular}
}
\end{table}

%% file: table/indicator_scaling.tex
\begin{table}[ht]
\centering
\caption{SR of Technical Indicator at Period $k$. \textbf{Bold} and \underline{Underline} means the best performance of the column across and within different blocks, respectively.}
\label{tab:indicators_scaling}
\resizebox{1.0\textwidth}{!}{%
\begin{tabular}{l|cc|cc|cc|cc|cc|cc|cc}
\toprule
  & \multicolumn{2}{c}{\textbf{CSI300}}
  & \multicolumn{2}{c}{\textbf{CSI500}}
  & \multicolumn{2}{c}{\textbf{NI225}}
  & \multicolumn{2}{c}{\textbf{SP500}}
  & \multicolumn{2}{c}{\textbf{FTSE100}}
  & \multicolumn{2}{c}{\textbf{TWSE300}}
  & \multicolumn{2}{c}{\textbf{Avg.}} \\
\textbf{Period $k$} & SR & CR & SR & CR & SR & CR & SR & CR & SR & CR & SR & CR & SR & CR \\
\midrule
3-only & 1.094 & 1.023 & 1.854 & 1.926 & 1.138 & 0.962 & \underline{\textbf{1.700}} & 3.694 & 0.589 & 0.385 & 1.189 & 1.196 & 1.261 & 1.531 \\
5-only & 1.229 & 1.183 & 1.868 & 1.708 & 1.014 & 0.902 & 1.555 & 3.407 & 0.532 & 0.357& 1.373 & 1.450 & 1.262 & 1.501 \\
10-only & \underline{1.333} & \underline{1.473} & 1.869 & 1.976 & 1.153 & 1.064 & 1.626 & 3.633 & \underline{\textbf{0.622}} & 0.403 & 1.382 & 1.519 & 1.331 & 1.678 \\
14-only & 1.243 & 1.194 & \underline{1.893} & 1.848 & \underline{\textbf{1.335}} & \underline{\textbf{1.236}} & 1.625 & \underline{3.773} & 0.551 & \underline{\textbf{0.419}} & \underline{1.395} & 1.475 & \underline{1.340} & 1.657 \\
20-only & 1.021 & 0.929 & 1.846 & \underline{2.006} & 1.157 & 0.998 & 1.581 & 3.748 & 0.507 & 0.348 & 1.394 & 1.525 & 1.251 & 1.592 \\
\midrule
Oracle & 1.333 & 1.473 & 1.893 & 2.006 & \underline{1.335} & \underline{1.236} & \underline{1.700} & 3.773& \underline{0.622} & \underline{0.419} & 1.395 & 1.525 & \textbf{\underline{1.402}} & 1.739 \\
All windows & \textbf{\underline{1.520}} & \textbf{\underline{1.640}} & \textbf{\underline{1.982}} & \textbf{\underline{2.207}} & 1.245 & 1.146 & 1.645 & \textbf{\underline{3.774}} & 0.568 & 0.413 & \textbf{\underline{1.418}} & \textbf{\underline{1.549}} & 1.396 & \textbf{\underline{1.790}} \\
\bottomrule
\end{tabular}%
}
\end{table}

%% file: table/market_diversity.tex
\begin{table}[ht]
\centering
\caption{CR by market and leave one pretraining market out. \textbf{Bold} and \underline{Underline} means the best performance of the column across and within different blocks, respectively.}
\resizebox{0.8\textwidth}{!}{%
\begin{tabular}{lccccccc}
\toprule
\textbf{Markets} 
  & \multicolumn{1}{c}{\textbf{CSI300}}
  & \multicolumn{1}{c}{\textbf{CSI500}}
  & \multicolumn{1}{c}{\textbf{NI225}}
  & \multicolumn{1}{c}{\textbf{SP500}} 
  & \multicolumn{1}{c}{\textbf{FTSE100}} 
  & \multicolumn{1}{c}{\textbf{TWSE300}} 
  & \multicolumn{1}{c}{\textbf{Avg.}} \\
\midrule
\textbf{\textit{One Market Only}} \\
CSI300-only & 1.041 & 1.459 & 0.805 & 3.488 & 0.221 & 1.385 & \underline{1.397} \\
CSI500-only & \underline{1.119} & \underline{1.625} & 0.755 & 3.112 & 0.272 & 1.232 & 1.352 \\
NI225-only & 0.482 & 1.013 & \underline{0.828} & 3.760 & \underline{0.310} & 1.287 & 1.280 \\
SP500-only & 0.714 & 1.330 & 0.714 & 3.374 & 0.247 & \underline{1.464} & 1.262 \\
FTSE100-only & 0.493 & 0.916 & 0.762 & 3.469 & 0.288 & 1.363 & 1.215 \\
TWSE300-only & 0.291 & 0.865 & 0.771 & \underline{4.050} & 0.253 & 1.357 & 1.264 \\
\midrule
\textbf{\textit{Remove One Market}} \\
w/o CSI300 & \underline{1.466} & 2.051 & 1.028 & \textbf{\underline{4.796}} & 0.361 & 1.458 & \textbf{\underline{1.860}} \\
w/o CSI500 & 1.084 & 1.789 & 0.965 & 3.527 & 0.394 & \textbf{\underline{1.565}} & 1.554 \\
w/o NI225 & 1.146 & 2.041 & 1.037 & 4.003 & \underline{0.407} & 1.446 & 1.680 \\
w/o SP500 & 1.197 & \textbf{\underline{2.400}} & 1.072 & 4.104 & 0.251 & 1.525 & 1.758 \\
w/o FTSE100 & 1.281 & 2.027 & \textbf{\underline{1.182}} & 3.659 & 0.388 & 1.498 & 1.672 \\
w/o TWSE300 & 1.208 & 2.056 & 1.070 & 3.882 & 0.339 & 1.390 & 1.658 \\
\midrule
\midrule
\textbf{\textit{Overall}} \\
Single Market & 1.041 & 1.626 & 0.828 & 3.374 & 0.288 & 1.357 & 1.419 \\
OOD performance & 1.466 & 1.789 & 1.037 & \underline{4.104} & 0.388 & 1.390 & 1.696 \\
Joint All Market & \textbf{\underline{1.640}} & \underline{2.207} & \underline{1.146} & 3.774 & \textbf{\underline{0.413}} & \underline{1.549} & \underline{1.790} \\
\bottomrule
\label{tab:analysis_market}
\end{tabular}
}
\end{table}

%% file: content/appendix_extra_analysis.tex
\section{Extra Analysis}
\label{app:extra_analysis}

\subsection{Preprocessing and Normalization Ablations}
\label{app:preprocess_ablation}
Table~\ref{tab:preprocess_ablation} evaluates the normalization stages introduced in Section~\ref{sec:targets} while keeping the remaining training configuration fixed.
At the sample level, the default max-scaling achieves the strongest performance (1.396/1.790 SR/CR). Replacing both feature and target scaling with the z-based alternative remains competitive (1.344/1.645), whereas mismatched transformations can degrade performance, particularly when z-normalized features are paired with max-normalized targets (0.890/0.818).
These results motivate applying compatible sample-level transformations to the input and scale-dependent indicator targets.

Per-market target standardization has a larger effect on optimization: removing it lowers average SR to 0.988, while removing IQR clipping alone has a smaller effect (1.296). Removing both operations substantially degrades performance to 0.601.
The cross-market balancing mechanisms provide more moderate improvements: performance remains at 1.316 without balanced market sampling and 1.345 without pooled re-standardization. Notably, both settings remain above single-market pretraining (1.183; Table~\ref{tab:indicator_analysis}), indicating that the cross-market gain is not solely explained by these balancing operations.

\input{table/preprocess_ablation}

We also vary the historical context available during pretraining. Increasing the context from 20 to 60 and 120 bars improves average SR/CR from 1.259/1.482 to 1.300/1.676 and 1.396/1.790, respectively. Note that this context length is distinct from the indicator period parameters analyzed in Section~\ref{sec:indicator_design} and from the fixed downstream input window.
Pretraining context length is not directly comparable across methods: the pretraining baselines use context length ranging from 20 to 64 bars, while $^\ddagger$Kronos and $^\dagger$FinCast are pretrained with 90 and 128 bars, respectively. 
Even at \model{}'s shortest setting of 20 bars, which matches the downstream input window, it already exceeds every pretraining baseline in Table~\ref{tab:main_results} and matches $^\ddagger$Kronos-small, indicating that its advantage does not stem from a longer pretraining context.

\subsection{Robustness to Downstream Transfer Protocol}
\label{app:main_with_freeze}
Different pretraining methods may exhibit different downstream adaptation behavior. To complement the transfer settings used in the main comparison, we re-evaluate all generic and financial pretraining methods under two shared downstream protocols: (i) frozen-encoder transfer, where only the common two-layer MLP prediction head is optimized, and (ii) full fine-tuning, where both the pretrained encoder and prediction head are updated. The downstream ranking objective, data splits, prediction head, and evaluation protocol are otherwise kept unchanged. We do not select between the two protocols on a per-method basis.
Table~\ref{tab:main_results_freeze} reports the corresponding market-wise results. Under frozen-encoder transfer, all three DIVINE backbones achieve higher average SR and CR than every compared pretraining baseline. The same pattern holds under full fine-tuning. The effect of adaptation itself is method-dependent—some pretrained encoders improve with full fine-tuning, whereas others perform better when frozen—but DIVINE remains strongest on average under either common protocol. These results show that the main performance advantage is robust to the choice of downstream adaptation strategy and is not explained by encoder freezing alone.
\input{table/main_result_freeze}

\subsection{Standard Deviations of Average Results}
\label{app:main_with_std}
Table~\ref{tab:main_results_std} complements Table~\ref{tab:main_results} by reporting the mean and standard deviation of the average SR and CR across the 25 seed--offset evaluations.
For each seed--offset combination, SR and CR are first averaged across the six markets, and the standard deviation is then computed over the resulting 25 average values.
\input{table/main_result_std}

\subsection{Main Results with Transaction Cost}
\label{app:main_with_cost}
We evaluate robustness to transaction costs by incorporating market-specific brokerage  commissions, exchange fees, and regulatory charges. 
Per-side (buy/sell) rates are approximately: CSI300 \& CSI500 0.006\%/0.056\%, NI225 0.002\%/0.002\%, S\&P500 negligible ($\sim$0\%), FTSE100 0.55\%/0.05\%, and TWSE300 0.1425\%/0.4425\%. 
At each rebalance, costs are charged on the actual change in portfolio weights rather than full turnover, making Table~\ref{tab:main_with_cost} a realistic rather than  worst-case estimate.

In Table~\ref{tab:main_with_cost}, we compare \model{} with the five strongest methods from Table~\ref{tab:main_results}.
Results show that \model{} maintains the highest average SR/CR of 1.067/1.483 under these conditions, demonstrating robustness to realistic trading frictions. The exception is FTSE100, where the relatively high per-side costs cause all methods to yield negative SR/CR, consistent with the already modest pre-cost performance observed in Table~\ref{tab:main_results}.

\input{table/main_with_cost}

\subsection{Analysis of IC and RankIC Metrics}
We report IC-based metrics in addition to our primary portfolio metrics to provide a more comprehensive evaluation.
Our primary evaluation follows standard practice in quantitative finance, where portfolio construction selects only the top-$k$ ranked stocks, making Sharpe Ratio (SR) and Calmar Ratio (CR) the more direct measures of practical model utility.
IC, by contrast, evaluates the quality of the \textit{entire} ranking distribution; a model with strong overall ranking correlation does not necessarily place the most profitable stocks at the top---a fundamental mismatch with the top-$k$ selection regime.

This discrepancy is evident when comparing cross-market averages across Tables~\ref{tab:main_results} and~\ref{tab:ic_results}: $^\ddagger$Kronos-small and PatchTST achieve average SR/CR of 1.256/1.404 and 1.007/1.202, respectively, yet PatchTST outperforms $^\ddagger$Kronos-small on IC-based metrics despite underperforming in portfolio returns. The same divergence appears within individual markets. 
On TWSE300, PatchTST attains the highest IC and RankIC while \model{} achieves the higher portfolio SR/CR (1.418/1.549 vs.\ 1.371/1.465), illustrating that stronger overall rank correlation need not translate into better top-$k$ selection. 
On FTSE100, the market where \model{} is weakest in portfolio terms (Section~\ref{sec:main_results}), IC-based metrics likewise rank \model{} below several baselines, consistent with the limitation discussed there.

Averaged across markets, \model{} nonetheless ranks first on both IC and RankIC, indicating that its portfolio advantage does not come at the cost of overall ranking quality. We report these metrics for completeness while noting that per-market divergences between the two metric families reinforce our use of SR and CR as the primary evaluation criteria.

\input{table/ic_table}

%% file: table/preprocess_ablation.tex
\begin{table}[ht]
\caption{Ablation of preprocessing and normalization choices.
All results report average performance across six markets.}
\label{tab:preprocess_ablation}
\centering
\resizebox{0.7\textwidth}{!}{%
\begin{tabular}{llcc}
\toprule
\textbf{Component} & \textbf{Variant} & \textbf{Avg. SR} & \textbf{Avg. CR} \\
\midrule
Default
& \model{} & 1.396 & 1.790 \\
\midrule

\multirow{3}{*}{Sample-level normalization}
& feat z-norm, target z-norm & 1.344 & 1.645 \\
& feat z-norm, target max-norm & 0.890 & 0.818 \\
& feat max-norm, target z-norm & 1.312 & 1.523 \\
\midrule

\multirow{3}{*}{Per-market target normalization}
& w/o standardization & 0.988 & 1.342 \\
& w/o IQR clipping & 1.296 & 1.751 \\
& w/o both & 0.601 & 0.603 \\
\midrule

\multirow{2}{*}{Cross-market training}
& w/o balanced market sampling & 1.316 & 1.692 \\
& w/o pooled re-standardization & 1.345 & 1.676 \\
\bottomrule
\end{tabular}
}
\end{table}

%% file: table/main_result_freeze.tex
\begin{table*}[ht]
\caption{
Robustness to downstream transfer protocol.
We report portfolio performance under frozen-encoder transfer and full fine-tuning across six markets.}

\centering
\setlength{\tabcolsep}{4pt}
\resizebox{\textwidth}{!}{%
\begin{tabular}{lc|cc|cc|cc|cc|cc|cc|cc}
\toprule
\textbf{Markets} $\rightarrow$ & & \multicolumn{2}{c}{\textbf{CSI300}} & \multicolumn{2}{c}{\textbf{CSI500}} & \multicolumn{2}{c}{\textbf{NI225}} & \multicolumn{2}{c}{\textbf{SP500}} & \multicolumn{2}{c}{\textbf{FTSE100}} & \multicolumn{2}{c}{\textbf{TWSE300}} & \multicolumn{2}{c}{\textbf{Avg.}} \\
\textbf{Method} & \textbf{Transfer} & SR & CR & SR & CR & SR & CR & SR & CR & SR & CR & SR & CR & SR & CR \\
\midrule

\multicolumn{16}{l}{\textit{\textbf{General Time-Series Pretraining}}} \\

TS2Vec & Frozen & 0.927 & 0.744 & 1.210 & 1.160 & 0.733 & 0.582 & 0.914 & 0.832 & 0.545 & 0.371 & 1.204 & 1.175 & 0.922 & 0.811 \\
& Full FT & 1.014 & 0.893 & 1.522 & 1.640 & 0.780 & 0.664 & 0.952 & 1.008 & 0.493 & 0.344 & 1.140 & 0.989 & 0.983 & 0.923 \\

PatchTST & Frozen & 0.584 & 0.407 & 0.859 & 0.648 & 0.744 & 0.616 & 1.580 & 2.901 & 0.533 & 0.369 & 0.945 & 0.829 & 0.874 & 0.962 \\
& Full FT & 0.662 & 0.533 & 1.147 & 1.113 & 0.807 & 0.653 & 1.618 & 3.117 & 0.437 & 0.334 & 1.371 & 1.465 & 1.007 & 1.202 \\

TimesURL & Frozen & 1.100 & 1.140 & 1.479 & 1.616 & 0.584 & 0.472 & 1.068 & 1.306 & 0.496 & 0.340 & 1.024 & 0.944 & 0.959 & 0.970 \\
& Full FT & 1.022 & 0.972 & 1.302 & 1.277 & 0.576 & 0.448 & 1.057 & 1.191 & 0.412 & 0.268 & 1.117 & 0.958 & 0.914 & 0.852 \\
\midrule

\multicolumn{16}{l}{\textit{\textbf{Financial Pretraining}}} \\

MERA & Frozen & 0.729 & 0.554 & 1.283 & 1.277 & 1.328 & 1.192 & 1.459 & 2.870 & 0.346 & 0.281 & 1.222 & 1.102 & 1.061 & 1.213 \\
& Full FT & 1.075 & 1.055 & 0.677 & 0.472 & 0.769 & 0.554 & 1.117 & 1.131 & 0.704 & 0.469 & 1.116 & 0.955 & 0.962 & 0.833 \\

CI-STHPAN & Frozen & 0.426 & 0.352 & 1.082 & 0.821 & 0.591 & 0.450 & 1.719 & 3.697 & 0.313 & 0.221 & 0.409 & 0.339 & 0.757 & 0.980 \\
& Full FT & 0.720 & 0.518 & 0.949 & 0.646 & 0.286 & 0.261 & 1.527 & 2.075 & 0.705 & 0.553 & 0.520 & 0.423 & 0.785 & 0.746 \\

FASCL & Frozen & 0.690 & 0.597 & 0.964 & 0.831 & 1.044 & 0.878 & 1.072 & 1.337 & 0.514 & 0.355 & 0.929 & 0.800 & 0.869 & 0.800 \\
& Full FT & 0.547 & 0.485 & 0.951 & 0.881 & 0.954 & 0.742 & 1.300 & 1.713 & 0.718 & 0.504 & 1.246 & 1.275 & 0.953 & 0.933 \\

SSPT & Frozen & 0.754 & 0.604 & 0.870 & 0.669 & 0.938 & 0.771 & 0.867 & 1.285 & 0.585 & 0.378 & 0.908 & 0.701 & 0.820 & 0.735 \\
& Full FT & 0.447 & 0.404 & 0.931 & 0.724 & 1.079 & 0.896 & 1.162 & 2.035 & 0.540 & 0.362 & 1.095 & 0.938 & 0.876 & 0.893 \\

\midrule
\multicolumn{16}{l}{\textit{\textbf{\model{} (Ours)}}} \\

GRU encoder & Frozen & 1.520 & 1.640 & 1.982 & 2.207 & 1.245 & 1.146 & 1.645 & 3.774 & 0.568 & 0.413 & 1.418 & 1.549 & 1.396 & 1.790 \\
& Full FT & 1.130 & 1.072 & 1.597 & 1.564 & 1.164 & 0.962 & 1.511 & 4.037 & 0.645 & 0.472 & 1.397 & 1.413 & 1.241 & 1.587 \\

Mamba encoder & Frozen & 1.083 & 1.183 & 1.675 & 1.841 & 1.075 & 0.961 & 1.807 & 3.519 & 0.645 & 0.425 & 1.273 & 1.336 & 1.259 & 1.544 \\
& Full FT & 1.344 & 1.385 & 1.704 & 1.513 & 1.034 & 0.913 & 1.071 & 3.066 & 0.666 & 0.568 & 1.406 & 1.434 & 1.204 & 1.479 \\

TFM encoder & Frozen & 1.045 & 0.939 & 1.512 & 1.498 & 1.170 & 0.989 & 1.516 & 3.144 & 0.323 & 0.161 & 0.994 & 1.039 & 1.093 & 1.295 \\
& Full FT & 1.383 & 1.371 & 1.811 & 1.813 & 1.193 & 0.970 & 1.456 & 3.586 & 0.646 & 0.446 & 1.510 & 1.835 & 1.333 & 1.670 \\

\bottomrule
\end{tabular}%
}
\label{tab:main_results_freeze}
\end{table*}

%% file: table/main_result_std.tex
\begin{table*}[ht]
\caption{Average portfolio performance with standard deviations across 25 seed--offset evaluations. Values report mean $\pm$ standard deviation for SR and CR averaged across the six equity markets.}
\centering
\newcommand{\std}[1]{{\scriptsize$\pm$#1}}
\resizebox{0.5\textwidth}{!}{%
\begin{tabular}{lc|cc}
\toprule
\textbf{Method} & \textbf{Param. (M)} & Avg. SR & Avg. CR \\

\midrule
\multicolumn{4}{l}{\textit{\textbf{Financial models trained from scratch}}} \\
RankLSTM & 0.05 & 0.768\,\std{0.216} & 0.772\,\std{0.135} \\
FinGAT & 0.04 & 0.764\,\std{0.129} & 0.595\,\std{0.148} \\
MASTER & 0.7 & 0.898\,\std{0.158} & 0.950\,\std{0.227} \\
AlphaMix & 0.04 & 0.873\,\std{0.112} & 0.931\,\std{0.191} \\
StockMixer & 0.01 & 0.647\,\std{0.099} & 0.518\,\std{0.102} \\
DHMoE & 10.2 & 0.902\,\std{0.166} & 0.901\,\std{0.149} \\

\midrule
\multicolumn{4}{l}{\textit{\textbf{General Time-Series Pretraining}}} \\
TS2Vec & 0.6 & 0.983\,\std{0.123} & 0.923\,\std{0.180} \\
PatchTST & 0.6 & 1.007\,\std{0.085} & 1.202\,\std{0.209} \\
TimesURL & 0.6 & 0.914\,\std{0.141} & 0.852\,\std{0.199} \\

\midrule
\multicolumn{4}{l}{\textit{\textbf{Financial Pretraining}}} \\
MERA & 1.5 & 0.962\,\std{0.127} & 0.833\,\std{0.209} \\
CI-STHPAN & 0.5 & 0.785\,\std{0.103} & 0.746\,\std{0.223} \\
FASCL & 14.2 & 0.953\,\std{0.146} & 0.933\,\std{0.190} \\
SSPT & 0.4 & 0.876\,\std{0.167} & 0.893\,\std{0.235} \\

\midrule
\multicolumn{4}{l}{\textit{\textbf{Financial Foundation Models}}} \\
FinCast & 991.4 & 0.754\,\std{0.104} & 0.583\,\std{0.112} \\
Kronos-mini & 5.3 & 1.110\,\std{0.109} & 1.142\,\std{0.198} \\
Kronos-small & 24.5 & 1.256\,\std{0.134} & 1.404\,\std{0.228} \\
Kronos-base & 99.7 & 1.154\,\std{0.162} & 1.228\,\std{0.252} \\

\midrule
\multicolumn{4}{l}{\textit{\textbf{\model{} (Ours)}}} \\
GRU encoder & 0.05 & 1.396\,\std{0.092} & 1.790\,\std{0.266} \\
Mamba encoder & 0.07 & 1.259\,\std{0.100} & 1.544\,\std{0.160} \\
TFM encoder & 0.1 & 1.333\,\std{0.096} & 1.670\,\std{0.302} \\

\bottomrule
\end{tabular}%
}
\label{tab:main_results_std}
\end{table*}

%% file: table/main_with_cost.tex
\begin{table*}[h]
\caption{Portfolio backtest results of best performing methods in Table~\ref{tab:main_results} with transaction cost. Best and second-best performance per column are highlighted in \textbf{bold} and \underline{underline}, respectively.}

\centering
\setlength{\tabcolsep}{4pt}
\resizebox{\textwidth}{!}{%
\begin{tabular}{l|cc|cc|cc|cc|cc|cc|cc}
\toprule
\textbf{Markets} $\rightarrow$ & \multicolumn{2}{c}{\textbf{CSI300}} & \multicolumn{2}{c}{\textbf{CSI500}} & \multicolumn{2}{c}{\textbf{NI225}} & \multicolumn{2}{c}{\textbf{SP500}} & \multicolumn{2}{c}{\textbf{FTSE100}} & \multicolumn{2}{c}{\textbf{TWSE300}} & \multicolumn{2}{c}{\textbf{Avg.}} \\
\textbf{Method} & SR & CR & SR & CR & SR & CR & SR & CR & SR & CR & SR & CR & SR & CR \\
\midrule
TS2Vec & 0.897 & 0.735 & 1.427 & 1.483 & 0.772 & 0.656 & 0.952 & 1.008 & -0.814 & -0.204 & 0.175 & 0.125 & 0.568 & 0.634 \\
PatchTST & 0.560 & 0.432 & 1.054 & 0.981 & 0.799 & 0.646 & \underline{1.618} & \underline{3.117} & -0.783 & -0.204 & \underline{0.598} & \underline{0.470} & 0.641 & 0.907 \\
MERA & 0.988 & 0.996 & 0.591 & 0.420 & 0.763 & 0.550 & 1.117 & 1.131 & \underline{-0.539} & \underline{-0.166} & 0.408 & 0.353 & 0.555 & 0.547 \\
FASCL & 0.457 & 0.395 & 0.873 & 0.785 & \underline{0.947} & 0.736 & 1.300 & 1.713 & \textbf{-0.471} & \textbf{-0.147} & 0.529 & 0.372 & 0.606 & 0.642 \\
Kronos-small & \underline{1.347} & \underline{1.342} & \underline{1.825} & \underline{2.061} & 0.895 & \underline{0.775} & 1.523 & 2.329 & -0.935 & -0.220 & 0.265 & 0.176 & \underline{0.820} & \underline{1.077} \\
\model{} & \textbf{1.424} & \textbf{1.487} & \textbf{1.904} & \textbf{2.070} & \textbf{1.239} & \textbf{1.139} & \textbf{1.645} & \textbf{3.774} & -0.541 & -0.168 & \textbf{0.733} & \textbf{0.595} & \textbf{1.067} & \textbf{1.483} \\
\bottomrule
\end{tabular}%
}
\label{tab:main_with_cost}
\end{table*}

%% file: table/ic_table.tex
\begin{table}[t]
\centering
\caption{Ranking-based performance across six markets. Best and second-best performance per column are highlighted in \textbf{bold} and \underline{underline}, respectively.}
\label{tab:ic_results}
\resizebox{0.6\textwidth}{!}{
\begin{tabular}{llcccc}
\toprule
\textbf{Market} & \textbf{Method} & \textbf{IC} & \textbf{ICIR} & \textbf{RankIC} & \textbf{RankICIR} \\
\midrule

\multirow{6}{*}{CSI300} & \model{} & \textbf{0.0503} & \textbf{0.3712} & \textbf{0.0437} & \textbf{0.3446} \\
 & MERA & 0.0323 & 0.2358 & 0.0298 & 0.2286 \\
 & PatchTST & \underline{0.0427} & \underline{0.3053} & \underline{0.0379} & 0.2930 \\
 & FASCL & 0.0218 & 0.1938 & 0.0188 & 0.1994 \\
 & TS2Vec & 0.0309 & 0.2871 & 0.0305 & \underline{0.2993} \\
 & Kronos-small & 0.0328 & 0.2737 & 0.0306 & 0.2738\\

\midrule
\multirow{6}{*}{CSI500} & \model{} & \textbf{0.0753} & \textbf{0.6191} & \textbf{0.0661} & 0.5631 \\
 & MERA & 0.0582 & 0.4815 & 0.0511 & 0.4551 \\
 & PatchTST & 0.0663 & 0.5364 & 0.0589 & 0.5128 \\
 & FASCL & 0.0591 & 0.5373 & 0.0459 & 0.5174 \\
 & TS2Vec & \underline{0.0677} & \underline{0.5968} & \underline{0.0595} & \textbf{0.6128} \\
 & Kronos-small & 0.0608 & 0.5495 & 0.0558 & \underline{0.5712}\\
 
\midrule
\multirow{6}{*}{NI225} & \model{} & \textbf{0.0138} & \textbf{0.1073} & \textbf{0.0130} & \textbf{0.1110} \\
 & MERA & \underline{0.0092} & 0.0815 & \underline{0.0096} & 0.0884 \\
 & PatchTST & 0.0043 & 0.0394 & 0.0072 & 0.0657 \\
 & FASCL & 0.0083 & \underline{0.0854} & 0.0084 & \underline{0.0975} \\
 & TS2Vec & 0.0060 & 0.0692 & 0.0076 & 0.0859 \\
 & Kronos-small & 0.0060 & 0.0667 & 0.0076 & 0.0832\\

\midrule
\multirow{6}{*}{S\&P500} & \model{} & \textbf{0.0352} & \underline{0.2392} & \textbf{0.0192} & \textbf{0.1476} \\
 & MERA & 0.0231 & 0.1592 & 0.0116 & 0.0792 \\
 & PatchTST & \underline{0.0275} & 0.2105 & \underline{0.0149} & \underline{0.1346} \\
 & FASCL & 0.0155 & 0.1317 & 0.0093 & 0.1185 \\
 & TS2Vec & 0.0032 & 0.0383 & 0.0050 & 0.0661 \\
 & Kronos-small & 0.0214 & \textbf{0.2417} & 0.0120 & 0.1277 \\
 
\midrule
\multirow{6}{*}{FTSE100} & \model{} & 0.0084 & 0.0529 & 0.0094 & 0.0632 \\
 & MERA & 0.0072 & 0.0477 & 0.0079 & 0.0542 \\
 & PatchTST & \underline{0.0098} & \underline{0.0692} & \underline{0.0099} & \underline{0.0723} \\
 & FASCL & \textbf{0.0143} & \textbf{0.1014} & \textbf{0.0129} & \textbf{0.0966} \\
 & TS2Vec & 0.0067 & 0.0527 & 0.0085 & 0.0672 \\
 & Kronos-small & 0.0075 & 0.0597 & 0.0083 & 0.0644 \\

\midrule
\multirow{6}{*}{TWSE300} & \model{} & \underline{0.0319} & \underline{0.2391} & 0.0402 & \underline{0.3426} \\
 & MERA & 0.0279 & 0.2128 & \underline{0.0406} & 0.3325 \\
 & PatchTST & \textbf{0.0324} & \textbf{0.2639} & \textbf{0.0424} & \textbf{0.3696} \\
 & FASCL & 0.0267 & 0.2168 & 0.0363 & 0.3403 \\
 & TS2Vec & 0.0203 & 0.2141 & 0.0255 & 0.2711 \\
 & Kronos-small & 0.0242 & 0.2208 & 0.0311 & 0.3055 \\

\midrule
\multirow{6}{*}{\textbf{Average}} & \model{} & \textbf{0.0358} & \textbf{0.2715} & \textbf{0.0319} & \textbf{0.2620} \\
 & MERA & 0.0263 & 0.2031 & 0.0251 & 0.2063 \\
 & PatchTST & \underline{0.0305} & \underline{0.2375} & \underline{0.0285} & \underline{0.2413} \\
 & FASCL & 0.0243 & 0.2111 & 0.0219 & 0.2283 \\
 & TS2Vec & 0.0225 & 0.2097 & 0.0228 & 0.2337 \\
 & Kronos-small & 0.0255 & 0.2354 & 0.0242 & 0.2376 \\

\bottomrule
\end{tabular}
}
\end{table}